\documentclass[journal]{IEEEtran}
\ifCLASSINFOpdf
\else
\fi
\usepackage{mathrsfs}
\usepackage{subfigure}
\usepackage{graphics}
\usepackage{multirow}
\usepackage{multicol}
\usepackage{amsmath}
\usepackage{booktabs}       % professional-quality tables
\usepackage{amsfonts}       % blackboard math symbols
\usepackage{nicefrac}       % compact symbols for 1/2, etc.
\usepackage{microtype}     % microtypography
\usepackage[pdftex]{graphicx}
\usepackage{bm}

\begin{document}
\title{Visual Space Optimization for Zero-shot Learning}

\author{Xinsheng Wang,
        Shanmin Pang,
        Jihua Zhu,
        Zhongyu Li,
        Zhiqiang Tian,
        and Yaochen Li% <-this % stops a space
\thanks{X. Wang, S. Pang, J. Zhu, Z. Li, Z. Tian, and Y. Li are with the School of Software Engineering, Xi'an Jiaotong University, Xi'an 710049, China. (email: wangxinsheng@stu.xjtu.edu.cn; pangsm@xjtu.edu.cn; zhujh@mail.xjtu.edu.cn; zhongyuli@xjtu.edu.cn; zhiqiangtian@mail.xjtu.edu.cn; yaochenli@mail.xjtu.edu.cn).}% <-this % stops a space
\thanks{Manuscript received xxxxx, 2019; revised xxxxx, 2019.}}


\markboth{Journal of \LaTeX\ Class Files,~Vol.~14, No.~8, June~2019}%
{Shell \MakeLowercase{\textit{et al.}}: Visual Space Optimization for Zero-shot Learning}



\maketitle

\begin{abstract}
Zero-shot learning, which aims to recognize new categories that are not included in the training set, has gained popularity owing to its potential ability in the real-word applications. Zero-shot learning models rely on learning an embedding space, where both semantic descriptions of classes and visual features of instances can be embedded for nearest neighbor search. Recently, most of the existing works consider the visual space formulated by deep visual features as an ideal choice of the embedding space. However, the discrete distribution of instances in the visual space makes the data structure unremarkable. We argue that optimizing the visual space is crucial as it allows semantic vectors to be embedded into the visual space more effectively. In this work, we propose two strategies to accomplish this purpose. One is the visual prototype based method, which learns a visual prototype for each visual class, so that, in the visual space, a class can be represented by a prototype feature instead of a series of discrete visual features. The other is to optimize the visual feature structure in an intermediate embedding space, and in this method we successfully devise a multilayer perceptron framework based algorithm that is able to learn the common intermediate embedding space and meanwhile to make the visual data structure more distinctive. Through extensive experimental evaluation on four benchmark datasets, we demonstrate that optimizing visual space is beneficial for zero-shot learning. Besides, the proposed prototype based method achieves the new state-of-the-art performance.
\end{abstract}


% \fi
\IEEEpeerreviewmaketitle

\section{Introduction}
\IEEEPARstart{I}{n} the past decade, the field of image recognition has been revolutionized by the emergence of learned deep representations \cite{Krizhevsky2012ImageNet,Simonyan2014Very,He2016Deep}. However, most of the popular recognition frameworks rely on a sufficient number of training samples, and the learned recognition algorithms only can be operated in a limited condition where the categories are included in the training set. In reality, training a particular model for each class is infeasible due to the insufficient training instances. On one hand, long-tailed distribution \cite{Salakhutdinov2011Learning,Zhu2014Capturing} arises in the frequencies of observing objects, that some popular categories have a large number of instances while some other categories have few or even no instances for training. On the other hand, new concepts are ever-growing, for which collecting and labeling sufficient large training sets for each category could be difficult and expensive. In these circumstances, training an effective classification system for newly appeared categories that are not included in the training set is necessary for the use of the learned model in the real-world applications.

\begin{figure}[t]
  \centering
  \includegraphics[width=\linewidth]{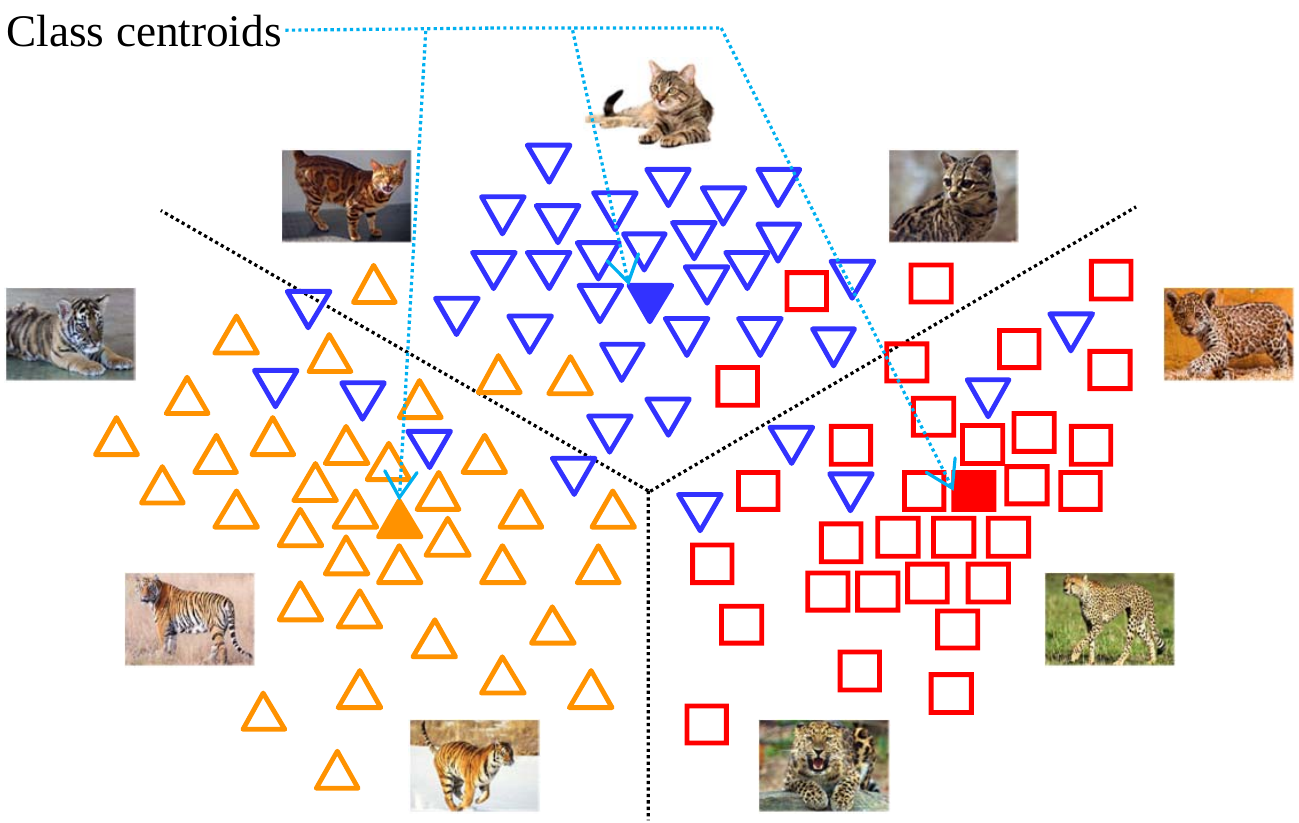}
  \caption{Illustration of visual feature distribution from three different categories, i.e. Cat, Tiger and Leopard. In some cases, the inter-class variation is even smaller than the intra-class. Even the class centroids are not discrimitive enough as they may be closer with instances from other classes than some from the same class.} 
  \label{fig:Vdist}
\end{figure}

Inspired by the learning mechanism of human on the recognition of new instances, zero-shot learning (ZSL) \cite{Lampert2009Learning,Larochelle2008Zero,Palatucci2009Zero} has been proposed and received a significant amount of interest. Humans are able to recognize new objects with the help of attribute descriptions and some related background knowledge. For instance, with knowledge of ``horse'' and ``black-and-white stripe'', when we are told that ``zebras are horse like animals united by their distinctive black-and-white striped coats'' we can recognize a zebra even if we never seen a zebra before. This is because we can associate the side information ``horse like'' and ``black-and-white stripe'' with zebras. Similarly, the key idea of ZSL is to capture the relationship between the knowledge contained in the seen and unseen instances with the help of side information which is also called auxiliary information.

Auxiliary information of ZSL is usually expressed in a high dimensional vector space called semantic space where seen and unseen classes are related. Class attribute vectors  \cite{Farhadi2009Describing,Farhadi2010Attribute,Vittorio2008Learning} and word vectors \cite{Mikolov2013Distributed,Lei2015Predicting,Elhoseiny2014Write} are most adopted as semantic representations in the semantic space. Given a set of visual features and semantic representations of the seen classes, the task of ZSL is to learn a joint embedding space where both visual features and semantic representations can be compared directly. With the learned projection functions, the visual features and semantic representations of the unseen test classes can be mapped into the embedding space, in which the recognition can be conducted by simple search of the nearest neighbor class prototype for each test instance.

Recent researches found that taking the visual space as the embedding space is favorable for the ZSL, because of its ability on alleviating the hubness problem \cite{Radovanovic2010Hubs,Shigeto2015Joint}. However, the visual instance features are discretely distributed in the visual space and each class contains numerous instance features. This means that the embedded semantic vector of one class should try to be closer to every visual instance features of the same class. The problem is that the visual features learned by CNNs are not always discriminative enough for discriminating the intra- or inter-class relationship. As illustrated in Fig. \ref{fig:Vdist}, the intra-class distance is sometimes even larger than the inter-class distance, and this significantly inhibits the learning of embedding functions.

% Even though there are lots of works  \cite{Annadani2018Preserving,Jiang2018Learning,Ye2017Zero,Chen2018Zero-shot} that have paid attention to keep the structure of semantic space or visual space, few works try to optimize the visual space structure. 

% To make the distribution of visual representations have a better structure after being projected into embedding space than that of original data structure in the visual space, 

Realizing this situation, this work proposes two methods, which try to learn the visual prototypes and to optimize the visual data structure respectively. Concretely, the visual prototype based method learns a class prototype for visual features of each class, thus an embedded semantic feature just need to be closer to its corresponding visual prototype rather than every visual feature of the same class. In addition, the visual prototype learned by the cross-entropy loss is more discriminative than the visual feature centroid of one class gotten by an average way. As for the second method in this work, we propose a flexible multilayer perceptron framework that not only maps both visual features and semantic representations into an intermediate embedding space, but also ensures a better embedded visual data structure. In this method, the network is trained with ranking loss and structure optimizing loss. Specifically, the ranking loss encourages matched image feature and attribute representation pairs have high similarities, while the structure optimizing loss is to make image pairs in the same category have smaller distance than those from different categories. To sum up, our contributions are:

\begin{itemize}
\item Propose a visual prototype based method for ZSL, in which the visual space is composed by visual feature prototypes instead of the visual instance features. With the cross-entropy loss, the proposed learnable visual prototypes are more discriminative than the visual centroids.
\item Propose a simple and effective visual space optimization framework for ZSL, which is able to optimize the distribution structure of visual features during the embedding process. Combined with the proposed structure optimizing function, two kinds of embedding loss, including simple ranking loss and bi-directional ranking loss are considered for ZSL.
\item Evaluate both of the proposed methods extensively on several popular datasets for ZSL, including AwA1 \cite{Lampert2009Learning}, AwA2 \cite{Xian2017Zero}, CUB  \cite{WelinderEtal2010} and SUN \cite{Xiao2010SUN}, and the results show that the proposed methods achieve the state-of-the-art performance. 
\end{itemize}

The rest of this paper is organized as follows: Section 2 covers the related work on zero-shot learning, embedding space and information preservation in the zero-shot learning. Section 3 describes the proposed approach in detail. Sections 4 and 5 present the experimental evaluation and related discussions respectively. Finally, the paper is concluded in Section 6.

\section{Related works}
In this section, we first give an overview of zero-shot learning, and then we briefly discuss the embedding space and data structure preservation in the zero-shot learning task.

\subsection{Zero-shot learning}
In the ZSL task, the seen categories in the training set and the unseen categories in the testing set are disjoint. In fact, ZSL can be seen as a subfield of transfer learning \cite{Pan2010A,Weiss2016A}, as the key idea of ZSL is to transfer the knowledge contained in the training resources to the task of testing instance classification. Early ZSL works \cite{Lampert2009Learning,Farhadi2009Describing,Lampert2014Attribute} follow an intuitive way to object recognition that makes use of the attributes to infer the label of an unseen test image. Recently, learning an embedding function that maps the semantic vectors and visual features into an embedding space, where the visual features and semantic vectors can be compared directly, shows outstanding performance and has been the most popular method \cite{Zhao2018Domain,Liu2018Zero,Annadani2018Preserving,Long2017Zero}. After the projection, nearest neighbor searching methods can be used to find the most similar class attribute vector for the test instance, and the discovered attribute corresponds to the most likely class. The embedding based method is adopted in this work.

Most recently, unseen class information is used to get better performance in the ZSL task  \cite{Jiang2018Learning,Liu2018Generalized,Verma2018Generalized,Mishra2018A,Xian2018Feature,Felix2018Multi,xian2019f}. For instance, in the work \cite{Jiang2018Learning}, unseen information is employed to assist aligning of the visual-semantic structures. As another example, some recent works \cite{Verma2018Generalized,Mishra2018A,Xian2018Feature,Felix2018Multi,xian2019f} adopt generative models to enlarge synthesized labeled examples from the unseen classes, and consequently, these examples can be assisted to train a better projection model. Furthermore, a related scenario is the transductive zero-shot learning \cite{Jie2018Transductive,Fu2015Transductive,Guo2016Transductive,Zhao2018Domain} , which assumes that the unlabeled samples from unseen classes are available during training. However, those works to some extent breach the strict ZSL settings that the testing resources should not be accessed in the training stage. In our work, we make no use of unseen classes information and consider only the seen resources are available at training time.

Compared with the strict ZSL, there is a more realistic and challenging task which is called generalized zero-shot learning (GZSL). Its targets include both seen and unseen categories. The problem of GZSL is proposed at the very beginning of ZSL work \cite{Lampert2009Learning}, and most of the above mentioned literatures evaluate their methods on both ZSL and GZSL settings. In this work, we also take GZSL into account. 

\begin{figure*}[htb]
\centering
\subfigure[Visual prototype based method]{
 \begin{minipage}[t]{\linewidth}
 \centering
 \includegraphics[scale = 0.6]{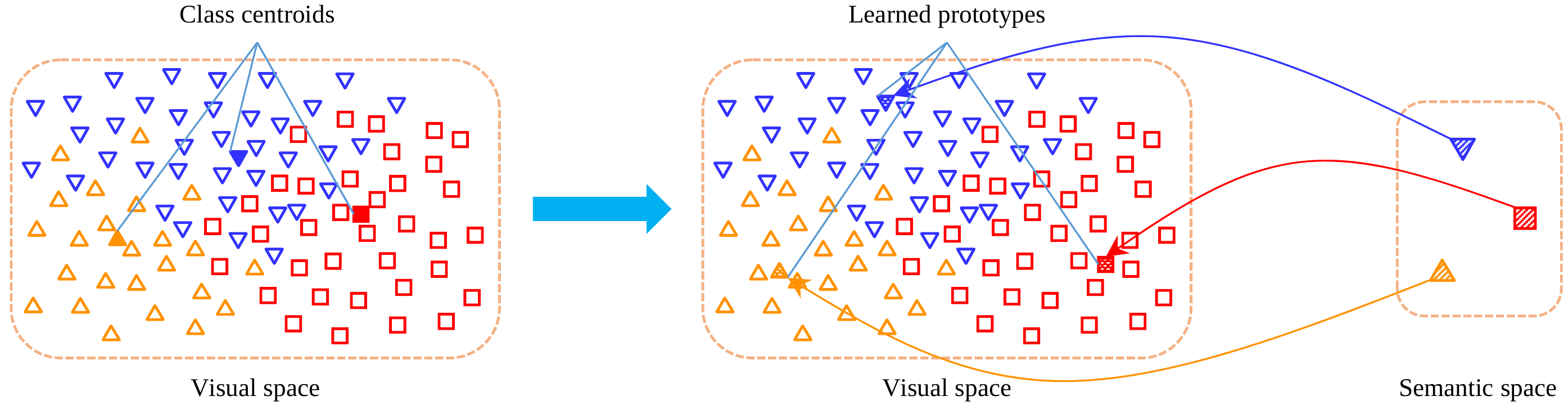}
 \label{fig:ill_a}
 \end{minipage}
 }\\
\subfigure[Visual feature structure optimization based method]{
  \begin{minipage}[t]{\linewidth}
  \includegraphics[scale = 0.6]{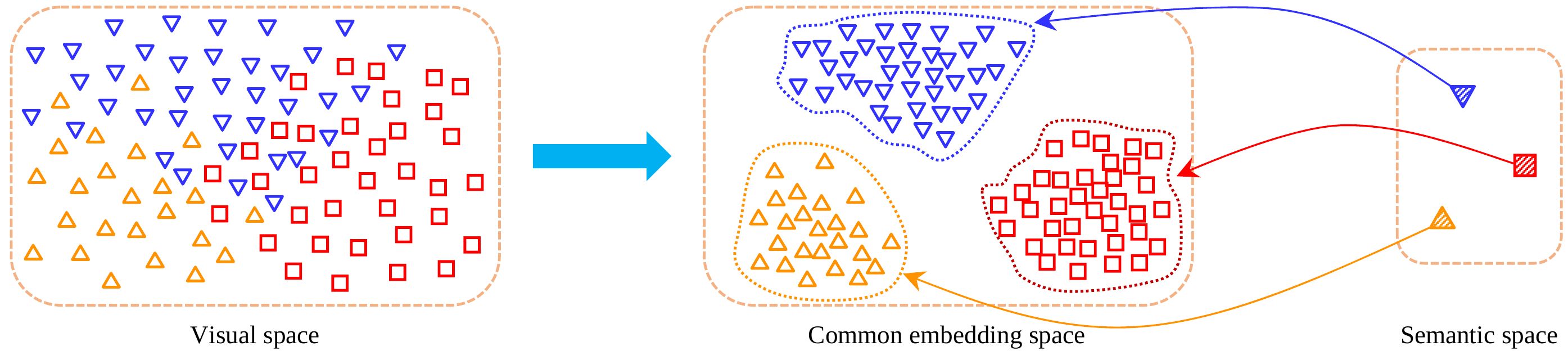}
  \centering
  \label{fig:ill_b}
  \end{minipage}
  }
 \centering
 \caption{Illustration of the proposed method. (a)Visual prototype based method. The prototypes are learned via backpropagation. With the learned visual prototypes, the semantic representation of each class is embedded to the corresponding visual prototype rather than numerous instance features. (b) Visual feature structure optimization based method. Both semantic representations and visual features are embedded into an intermediate space. The dimensions in the embedding space are same as those in visual space.}
 \label{fig:ill_proposed_methos}
 \end{figure*}

\subsection{Embedding space}
The choice of the embedding space is a key to the success of a ZSL model. Semantic space is often chosen as the embedding space in lots of researches  \cite{Xian2016Latent,Romera-Paredes2015An,Akata2015Evaluation,Frome2013Devise}. Owing to the advantage that each class is represented by one semantic vector in the semantic space, taking the semantic space as the embeddding space is helpful for the  better embedded visual data structure. However, on the downside, this strategy will significantly shrink the variance of the data points and thus aggravate the hubness problem \cite{Radovanovic2010Hubs,Shigeto2015Joint}. To alleviate this problem, some recent works \cite{Shigeto2015Joint,Zhang2017Learning} choose the visual space as the embedding space and map the semantic vectors to the visual space. However, using the visual space as embedding space faces a new problem. Instance features in the visual space are not distributed in an ideal structure due to the possibility of large inter-class similarities and small intra-class similarities. 

Common intermediate embedding space is also popular in the literature \cite{Changpinyo2016Synthesized,Zhang2015Zero}. Besides, more than one projection method can be realized in some works \cite{Jiang2018Learning,Kodirov2017Semantic,Zhao2018Domain} in the testing process. For instance, in the work \cite{Jiang2018Learning}, an intermediate aligned space is learned using the class prototypes, and the recognition can be conducted in all three space, namely the visual space, the semantic space and the intermediate space. 

In those embedding strategies, the intermediate embedding space makes it possible to adjust data structures both of semantic vectors and visual features. Thus, the intermediate embedding space strategy is adopted in the proposed visual space optimization based method. Considering the intrinsic superiority of using the visual space as embedding space on alleviating the hubness problem \cite{Zhang2017Learning}, the intermediate space in this method is closer to the visual space instead of being equivalent to visual space and semantic space. Besides, in order to take the visual space as the embedding space with more discriminative structure, the other method proposed in this work is to learn the visual feature prototypes, so that each visual class can be represented by one visual prototype instead of numerous discrete visual features. 

\subsection{Structure preservation}
Since there is a huge gap between visual and semantic spaces, the learned model tends to not discover the intrinsic topological structure when maps the data into the embedding space. Some works \cite{Annadani2018Preserving,Jiang2018Learning,Kodirov2017Semantic,Zhao2018Domain,Morgado2017Semantically,Meng2018Meng,Zhang2016Zero-r,Li2017Zero,Deutsch2017Zero,Liu2018Zero,Zhang2018Triple} have been conducted to keep the data structure during the projection. Manifold learning is a popular method used to keep the data structure in the ZSL \cite{Morgado2017Semantically,Meng2018Meng,Zhang2016Zero-r,Li2017Zero,Deutsch2017Zero}. Taking the visual space as the embedding space, the work  \cite{Long2017Zero} introduces an auxiliary latent-embedding space with manifold regularization to reconcile the semantic space with the visual feature space, which can preserve the intrinsic data structural information of both visual and semantic spaces. 

Encoder-decoder paradigm has been taken to preserve the data structure in recent works  \cite{Annadani2018Preserving,Kodirov2017Semantic,Zhao2018Domain,Liu2018Zero,Zhang2018Triple}. In SAE \cite{Kodirov2017Semantic}, the encoder is used to learn a projection from the feature space to a semantic space and the decoder tries to reconstruct the original visual features. During the test, the unseen visual features can be projected to the semantic space by the encoder, or reverse projection can be realized by the decoder. Based on this work, LESAE \cite{Liu2018Zero} adds the low-rank constraint for the learned embedding space in the encoder and get better performance. DIPL \cite{Zhao2018Domain} extends the encoder-decoder method in a transductive learning ZSL task. The work of \cite{Annadani2018Preserving} conducts the encoder-decoder process with a multilayer perceptron framework, and three class relations, namely same class, semantically similar class and semantically dissimilar class based on the semantic similarity are considered to preserve the semantic vector structure in the embedding space. 

However, most existing ZSL methods, which put much attention on keeping the original visual structure, neglect  indistinguishable distribution of visual features. In this work, we are not to preserve the original visual feature structure like previous works but to optimize it. As illustrated in Fig. \ref{fig:ill_proposed_methos}, we propose two strategies to address the indistinguishable distribution of features in the visual space. One is to learn the visual prototypes with which one class in the visual space can be represented by one visual prototype feature rather than discrete instance features. The other is to optimize the visual data structure that makes the distance of embedding visual feature pairs in same class closer and make instances from inter classes have obvious boundaries.

% However, most existing ZSL methods neglect the intrinsic distribution of visual data that its original structure is not good enough as shown in Fig. 1, and put much attention on keeping the original visual structure. 
\section{ZSL with visual prototypes}

\subsection{Problem definition}
Let ${Y_s} = \left[ {y_1^{(s)},y_2^{(s)},\ldots,y_p^{(s)}} \right]$ denote a set of $k$-dimension semantic representations of $p$ seen classes $\mathcal{S} = \left\{ {{s_1},{s_2},\ldots,{s_p}} \right\}$, and ${Y_u} = \left[ {y_1^{(u)},y_2^{(u)},\ldots,y_q^{(u)}}\right]$ denote semantic vectors of $q$ unseen classes $\mathcal {U} = \left\{ {{u_1},{u_2},\dots,{u_q}} \right\}$. The seen and unseen classes are disjoint, i.e. $\mathcal{S} \cap \mathcal{U} = \phi$. $x_i^{( s )}$ is a $d$-dimension image feature from one seen class. The training set is given as ${\mathcal {D}_{tr}} = \left\{ {x_i^{\left( s \right)},y_{l{s_i}}^{\left( s \right)},i = 1,2,...,{N_s}} \right\}$, where $l{s_i} \in \left\{ {1,2,\dots,p} \right\}$ is the label of $x_i^{( s )}$ according to $\mathcal{S}$, $y_{l{s_i}}^{(s)}$ is the semantic vector of the $i$-th image, and ${N_s}$ denotes the total sample number in the training set. Similarly, the testing set with total sample number of ${N_u}$ is given as ${\mathcal{D}_{te}} = \left\{ {x_i^{\left( u \right)},y_{l{u_i}}^{\left( u \right)},i = 1,2...,{N_u}} \right\}$, where $x_i^{( u )} \in {\mathbb{R}^{d \times 1}}$ is a visual vector of the $i$-th image in the testing set, and $y_{l{u_i}}^{( u )} \in {\mathbb{R}^{k \times 1}}$ is the corresponding unseen semantic representation with the label $l{u_i} \in \left\{ {1,2,\dots,q} \right\}$. Given a new sample from the unseen class, the goal of the ZSL is to predict the correct class of the given sample with a learned model which is trained only with samples from seen classes.  
\subsection{Learning visual prototypes}
In the light of that each class in the visual space is composed by numerous instance features, we tend to use a visual prototype to represent visual features of one class. Intuitively, the visual feature centroids obtained by averaging visual features of each class can be adopted directly to work as visual prototypes. However, due to the defective instance feature distribution, as shown in Fig. \ref{fig:Vdist}, the centroid of features average is also not discriminative enough and it may have small distance with several instances from other classes. Therefore, we propose a learnable strategy to learn the visual prototypes via backpropagation. The visual prototypes are denoted as $z_i$ where $i \in \left\{ {1,2..,p} \right\}$ represents the index of the classes.

We take the visual prototypes learning process as a prototype-based classification problem. The difference is that we take nothing for the visual features but only update the prototypes themselves. Given an visual feature $x_i$, the similarity of the visual feature with the visual prototype $z_j$ can be denoted as:
\begin{equation}
    {h_{i,j}} = sim\left( {{x_i},{z_j}} \right)
\end{equation}
where $sim\left( { \cdot , \cdot } \right)$ is a similarity function, such as consine similarity and inner product, and the latter is adopted in this work. Then, we use the Softmax to get the final prediction confidence:
\begin{equation}
    {\hat h_{i,j}} = \frac{{\exp \left( {{h_{i,j}}} \right)}}{{\sum\limits_{k = 1}^p {\exp \left( {{h_{i,k}}} \right)} }}
\end{equation}

With the prediction confidences and corresponding labels, we can train the visual prototypes using the common cross-entropy loss, defined as:
\begin{equation}
    {{\cal L}_{proto}} =  - \sum\limits_{i = 1}^{{N_s}} {\sum\limits_{j = 1}^p {{s_{i,j}}\log \left( {{{\hat h}_{i,j}}} \right)} } 
\label{eq:loss_proto}
\end{equation}
where $s_{i,j}$ is an indicator function for label $x_{i}$ (i.e., one hot encoded vector).  It is worth noting that, compared with the traditional classifier training process, we update the visual prototypes rather than the visual features or any networks that process visual features.

With the learned visual prototypes, the semantic vectors can be projected into the corresponding visual prototypes in the visual space via a multilayer perceptron.

\subsection{Embedding the semantic representations}
With the learned visual prototypes, we only need to make the embedded semantic vectors close to their corresponding visual prototypes when embedding the semantic representations into the visual space. Thus, the object function for the embedding can be:
\begin{equation}
    \mathop {\min }\limits_\psi  \sum\limits_{i = 1}^p {{{\left\| {\psi \left( {y_i^{\left( s \right)}} \right) - {z_i}} \right\|}^2}}
\end{equation}
where $\psi ( . )$ is the embedding function of semantic vectors. In this work, we adopt a multilayer perceptron network to work as the embedding function, and the loss function can be written as:
\begin{equation}
\begin{array}{l}
\begin{aligned}
{{\cal L}_{emb}} = \sum\limits_{i = 1}^p {{{\left\| {f\left( {{W_2}f\left( {{W_1}y_i^{\left( s \right)}} \right) - {z_i}} \right)} \right\|}^2}} \\
{\rm{ + }}{\lambda _{emb}}\left( {{{\left\| {{W_1}} \right\|}^2} + {{\left\| {{W_2}} \right\|}^2}} \right)
\end{aligned}
\end{array}
\label{eq:loss_proto_emb}
\end{equation}
where ${W_1} \in {\mathbb{R}^{{M}\times k}}$ and ${W_2} \in {\mathbb{R}^{ d\times {M}}}$ are the first and second FC layer respectively. ${M}$ is the dimension of the hidden layer. $f( . )$ denotes the ReLU algorithm. ${\lambda _{emb}}$ is the hyperparameter weighting the parameters regularization losses.

\subsection{Recognition}
When the semantic embedding function $\psi ( . )$ is learned in the training stage, the recognition for the testing set can be realized. Given an image $x_i^{( u )}$ from the testing set, the recognition is achieved by finding the unseen class label ${u_j} \in \mathcal { U}$ according to its semantic vector $y_j^{( u )}$:
\begin{equation}
y_j^{\left( u \right)} = \mathop {{\mathop{\rm argmin}\nolimits} }\limits_{y_j^{\left( u \right)} \in {Y_u}} {\left\| {x_i^{\left( u \right)} - \psi \left( {y_j^{\left( u \right)}} \right)} \right\|^2}
\end{equation}

\section{ZSL with visual data structure optimization}
\subsection{Network architecture}
The aim of this method is to embed the visual features and semantic representations into a common intermediate embedding space, and meanwhile to optimize the structure of visual data. Thus, two kinds of loss functions are included in this part. One is the embedding loss that works to make the matched pairs of visual features and semantic vectors be closer. The other is the structure optimizing loss for optimizing the visual data structure. For the embedding loss, we consider two specific loss functions: simple ranking loss and bi-directional ranking loss.

\begin{figure*}[htb]
\centering
\subfigure[]{
 \begin{minipage}[t]{0.4\linewidth}
 \centering
 \includegraphics[scale = 0.6]{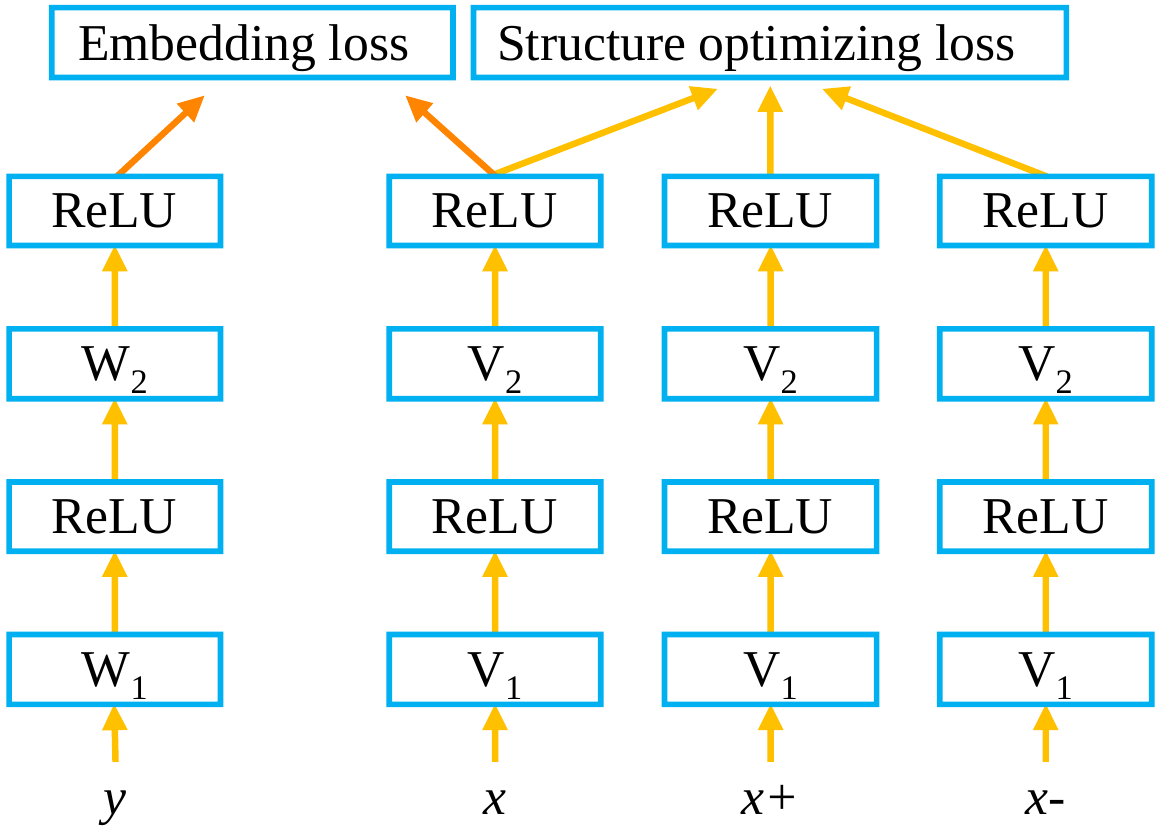}
 \label{fig:NetArchit_a}
 \end{minipage}
 }
\hspace{0.1in}
\subfigure[]{
  \begin{minipage}[t]{0.5\linewidth}
  \includegraphics[scale = 0.6]{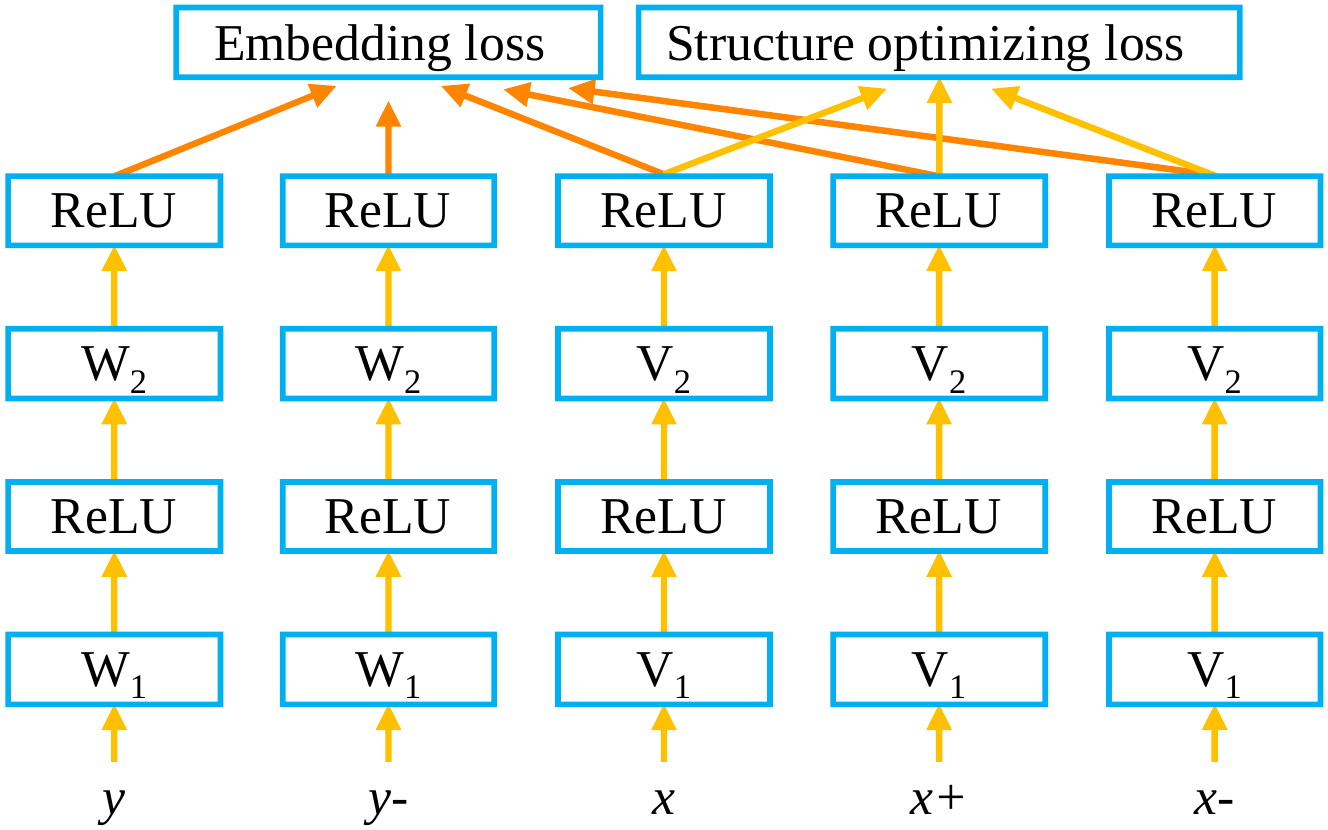}
  \centering
  \label{fig:NetArchit_b}
  \end{minipage}
  }
 \centering
 \caption{Proposed network architectures for visual feature structure optimization based methods. (a) The architecture with the simple ranking loss and the structure optimizing loss (SRS). (b) The architecture with the bi-directional ranking loss and structure optimizing loss (BRS).}
 \end{figure*}

According to the adopted specific embedding loss function, we create two different network architectures, as shown in Fig. \ref{fig:NetArchit_a} and Fig. \ref{fig:NetArchit_b}. These two networks share the same architecture of visual network branch and semantic network branch respectively. Both the semantic and visual embeddings are achieved by a multilayer perceptron framework which is same as that in the prototype based method for semantic embedding. Specifically, the multilayer perceptron takes a $k$-dimension semantic vector or a $d$-dimension visual representation as input, and after going through two fully connected (FC) layers + Rectified Linear Unit (ReLU) layers, it outputs a $L$-dimension embedding vector.

\subsection{Embedding loss}
We consider two kinds of functions for the embedding loss. One is called simple ranking loss and the other is the bi-directional ranking loss.
\subsubsection{ Simple ranking loss.}
Given a training set with matched pairs of $x_i^{( s )}$ and $y_{l{s_i}}^{( s )}$, the object function is:
\begin{equation}
    \mathop {\min }\limits_{\varphi ,\psi } \sum\limits_{i = 1}^{{N_s}} {{{\left\| {\varphi \left( {x_i^{\left( s \right)}} \right) - \psi \left( {y_{l{s_i}}^{\left( s \right)}} \right)} \right\|}^2}} 
\end{equation}
where $\varphi ( . )$ is the embedding function of visual features and $\psi ( . )$ is the embedding function of semantic vectors. According to Fig. \ref{fig:NetArchit_a}, the simple ranking loss function for the objection function is as follows:
\begin{equation}
    {\mathcal{ L}_{embs}} = \sum\limits_{i = 1}^{{N_s}} {{{\left\| {f\left( {{V_2}f\left( {{V_1}x_i^{\left( s \right)}} \right)} \right) - f\left( {{W_2}f\left( {{W_1}y_{l{s_i}}^{\left( s \right)}} \right)} \right)} \right\|}^2}}
    \label{eq:Lembs}
\end{equation}
where ${V_1} \in {\mathbb{R}^{{M_1} \times d}}$ and ${V_2} \in {\mathbb{R}^{ L\times {M_1}}}$ are the first and second FC layer of the visual embedding branch. Same as in the Eq. \ref{eq:loss_proto_emb}, ${W_1} \in {\mathbb{R}^{{M_2}\times k}}$ and ${W_2} \in {\mathbb{R}^{ L\times {M_2}}}$ are FC layers of the semantic embedding branch. $L$ is the dimension in the embedding space. ${M_1}$ and ${M_2}$ are dimensions of corresponding hidden layers. $f( . )$ denotes the ReLU algorithm.

\subsubsection{Bi-directional ranking loss.} Given a visual feature $x_i^{( s )}$, let $Y_i^{( s ) - }$ denote its set of non-matched (negative) semantic vectors. If $y_{l{s_i}}^{( s )}$ and $y_{l{s_k}}^{( s )}$ are positive and negative semantic vectors for $x_i^{( s )}$ respectively, the distance between the $x_i^{( s )}$ and $y_{l{s_i}}^{( s )}$ should be smaller than the distance between $x_i^{( s )}$ and $y_{l{s_k}}^{( s )}$  with a margin of ${m_1}$. Thus we can get a triplet-wise constraint, as follows:
\begin{equation}
    \begin{array}{*{20}{c}}
{{{\left\| {\varphi \left( {x_i^{\left( s \right)}} \right) - \psi \left( {y_{l{s_i}}^{\left( s \right)}} \right)} \right\|}^2} + {m_1} < {{\left\| {\varphi \left( {x_i^{\left( s \right)}} \right) - \psi \left( {y_{l{s_k}}^{\left( s \right)}} \right)} \right\|}^2}}\\
{\forall y_{l{s_k}}^{\left( s \right)} \in Y_i^{\left( s \right) - }}
\end{array}
\end{equation}
Similarly, given a semantic vector $y_{l{s_{i'}}}^{( s )}$, the analogous constraints in the other direction can be written as: 
\begin{equation}
    \begin{array}{*{20}{c}}
{{{\left\| {\psi \left( {y_{l{s_{i'}}}^{\left( s \right)}} \right) - \varphi \left( {x_{j'}^{\left( s \right)}} \right)} \right\|}^2} + {m_2} < {{\left\| {\psi \left( {y_{l{s_{i'}}}^{\left( s \right)}} \right) - \varphi \left( {x_{k'}^{\left( s \right)}} \right)} \right\|}^2}}\\
{\forall x_{j'}^{\left( s \right)} \in X_{i'}^{\left( s \right) + },\;\forall x_{k'}^{\left( s \right)} \in X_{i'}^{\left( s \right) - }}
\end{array}
\end{equation}
where $X_{i'}^{( s ) + }$ and $X_{i'}^{( s ) - }$ denote the sets of matched (positive) and non-matched (negative) visual features respectively for $y_{l{s_{i'}}}^{( s )}$.

These two constraints can be converted into a margin-based bi-directional ranking loss function:
\begin{equation}
\begin{array}{l}
\begin{aligned}
{{\mathcal L}_{embb}} = \sum\limits_{i,j,k} {\left[ {{m_1} + {{\left\| {\varphi \left( {x_i^{\left( s \right)}} \right) - \psi \left( {y_{l{s_i}}^{\left( s \right)}} \right)} \right\|}^2}} \right.} \\
{\left. { - {{\left\| {\varphi \left( {x_i^{\left( s \right)}} \right) - \psi \left( {y_{l{s_k}}^{\left( s \right)}} \right)} \right\|}^2}} \right]_ + }\\
 + \beta \sum\limits_{i',j',k'} {\left[ {{m_2} + {{\left\| {\psi \left( {y_{l{s_{i'}}}^{\left( s \right)}} \right) - \varphi \left( {x_{j'}^{\left( s \right)}} \right)} \right\|}^2}} \right.} \\
{\left. { - {{\left\| {\psi \left( {y_{l{s_{i'}}}^{\left( s \right)}} \right) - \varphi \left( {x_{k'}^{\left( s \right)}} \right)} \right\|}^2}} \right]_{\rm{ + }}}
\end{aligned}
\end{array}
\label{eq:lambb}
\end{equation}
where ${[ t ]_ + } = \max ( {t,0} )$, $\beta $ is the balance weight for the different directions. The scale of distance between an embedded visual feature and an embedded semantic feature has a huge change during the training, regardless of whether they come from the same class or not. With this in mind, we take self-adaptive margins instead of the fixed value in this work. Specifically, the margins $m_1$ and $m_2$ are computed by:
\begin{equation}
\begin{array}{l}
\begin{aligned}
{m_1} = {\alpha _1}{\left\| {\varphi \left( {x_i^{\left( s \right)}} \right) - \psi \left( {y_{l{s_i}}^{\left( s \right)}} \right)} \right\|^2}\\
 + {\alpha _1}{\left\| {\varphi \left( {x_i^{\left( s \right)}} \right) - \psi \left( {y_{l{s_k}}^{\left( s \right)}} \right)} \right\|^2}
 \end{aligned}
\end{array}
\end{equation}
\begin{equation}
\begin{array}{l}
\begin{aligned}
{m_2} = {\alpha _2}{\left\| {\psi \left( {y_{l{s_{i'}}}^{\left( s \right)}} \right) - \varphi \left( {x_{j'}^{\left( s \right)}} \right)} \right\|^2}\\
 + {\alpha _2}{\left\| {\psi \left( {y_{l{s_{i'}}}^{\left( s \right)}} \right) - \varphi \left( {x_{k'}^{\left( s \right)}} \right)} \right\|^2}
\end{aligned}
\end{array}
\end{equation}
where ${\alpha _1}$ and ${\alpha _2}$ are hyperparameters adjusting the value of the margins. 

\subsection{Visual data structure optimizing loss}
As shown in Fig. \ref{fig:Vdist}, the distribution of instances in the visual space tends to be indistinctive, thus we propose a structure optimizing constraint for the embedding of visual features to optimize the visual data structure. Let $N( {x_i^{( s )}} )$ denote the neighborhood of $x_i^{( s )}$ , which is a set of the visual features from the same class of $x_i^{( s )}$ . The purpose of the structure optimizing constraint is to enforce the distances between $x_i^{( s )}$ and neighborhood points and those outside of the neighborhood satisfy:
\begin{equation}
    \begin{array}{*{20}{c}}
{{{\left\| {\varphi \left( {x_i^{\left( s \right)}} \right) - \varphi \left( {x_j^{\left( s \right)}} \right)} \right\|}^2} + {m_3} < {{\left\| {\varphi \left( {x_i^{\left( s \right)}} \right) - \varphi \left( {x_k^{\left( s \right)}} \right)} \right\|}^2}}\\
{\forall x_j^{\left( s \right)} \in N\left( {x_i^{\left( s \right)}} \right),\;\forall x_k^{\left( s \right)} \notin N\left( {x_i^{\left( s \right)}} \right)}
\end{array}
\end{equation}
The corresponding loss function is described as:
\begin{equation}
\begin{array}{l}
\begin{aligned}
{\mathcal {L}_{opts}} = \sum\limits_{i,j,k} {\left[ {{m_3} + {{\left\| {\varphi \left( {x_i^{\left( s \right)}} \right) - \varphi \left( {x_j^{\left( s \right)}} \right)} \right\|}^2}} \right.} \\
{\left. { - {{\left\| {\varphi \left( {x_i^{\left( s \right)}} \right) - \varphi \left( {x_k^{\left( s \right)}} \right)} \right\|}^2}} \right]_{\rm{ + }}}
\end{aligned}
\end{array}
\label{eq:opts}
\end{equation}
where the margin is also defined as a self-adaptive value:
\begin{equation}
\begin{array}{l}
\begin{aligned}
{m_3} = {\alpha _3}{\left\| {\varphi \left( {x_i^{\left( s \right)}} \right) - \varphi \left( {x_j^{\left( s \right)}} \right)} \right\|^2}\\
 + {\alpha _3}{\left\| {\varphi \left( {x_i^{\left( s \right)}} \right) - \varphi \left( {x_k^{\left( s \right)}} \right)} \right\|^2}
 \end{aligned}
\end{array}
\end{equation}
According to the embedding loss whether is the simple ranking loss or the bi-directional ranking loss, the whole loss function can be written as two forms:
\begin{equation}
\begin{array}{l}
\begin{aligned}
{\mathcal {L}_{SRS}} = {\mathcal {L}_{embs}} + {\lambda _1}{\mathcal { L}_{opts}} + {\lambda _{embs1}}\left( {{{\left\| {{W_1}} \right\|}^2} + {{\left\| {{W_2}} \right\|}^2}} \right)\\
 + {\lambda _{embs2}}\left( {{{\left\| {{V_1}} \right\|}^2} + {{\left\| {{V_2}} \right\|}^2}} \right)
 \end{aligned}
\end{array}
\label{eq:Lsrs}
\end{equation}
\begin{equation}
\begin{array}{l}
\begin{aligned}
{\mathcal{ L}_{BRS}} = {\mathcal{ L}_{embb}} + {\lambda _2}{\mathcal{ L}_{opts}} + {\lambda _{embb1}}\left( {{{\left\| {{W_1}} \right\|}^2} + {{\left\| {{W_2}} \right\|}^2}} \right)\\
 + {\lambda _{embb2}}\left( {{{\left\| {{V_1}} \right\|}^2} + {{\left\| {{V_2}} \right\|}^2}} \right)
 \end{aligned}
\end{array}
\label{eq1:Lbrs}
\end{equation}
where ${\lambda _1}$ and ${\lambda _2}$ are hyperparameters weighting the strengths of the structure optimizing loss against the embedding loss. ${\lambda _{embs1}}$, ${\lambda _{embs2}}$, ${\lambda _{embb1}}$ and ${\lambda _{embb2}}$ are hyperparameters weighting the parameters regularization losses.
 
\subsection{Mining tuples}
The proposed algorithm relies on mining appropriate tuples for the training. Given a visual feature $x_i^{( s )}$ as the anchor sample, the corresponding semantic vector $y_{l{s_i}}^{( s )}$ is needed to compose the matched pair $\left( {x_i^{( s )},y_{l{s_i}}^{( s )}} \right)$ for the input of ${\mathcal { L}_{embs}}$ in Eq. (\ref{eq:Lembs}). Besides, tuples which contain the positive samples and negative samples are needed to optimize the visual data structure. In the tuple $\left( {x_i^{( s )},x_j^{( s )},x_k^{( s )}} \right)$ , the positive visual sample $x_j^{( s )}$ is chosen at random from the same class of the anchor sample $x_i^{( s )}$. The choosing of the negative samples plays an import role for the convergence of the training. In this work, we sample the negative samples in an online fashion, wherein for each iteration a criterion is evaluated, and the hardest negative $x_k^{( s )}$  for each anchor visual feature $x_i^{( s )}$ is sampled within a batch. With the input of tuples $\left( {y_{l{s_i}}^{( s )},x_i^{( s )},x_j^{( s )},x_k^{( s )}} \right)$, the network with the loss function ${\mathcal { L}_{SRS}}$ can be trained with Stochastic Gradient Descent (SGD). To optimize the loss function ${\mathcal { L}_{BRS}}$, the extra negative semantic vectors should be sampled. Because of the limited total number of semantic vectors, the hardest negative semantic vector $y_{l{s_k}}^{( s )}$ is sampled within all semantic vectors instead of within a batch. Then, the tuple sampled for ${\mathcal { L}_{BRS}}$ can be given as $\left( {y_{l{s_i}}^{( s )},y_{l{s_k}}^{( s )},x_i^{( s )},x_j^{( s )},x_k^{( s )}} \right)$.

\subsection{Recognition}
Similar with the recognition process in the prototype based method, with the learned visual embedding function $\varphi ( . )$ and semantic embedding function $\psi ( . )$ and testing image $x_i^{( u )}$, the recognition is achieved by finding the unseen class label ${u_j} \in \mathcal { U}$ according to its semantic vector $y_j^{( u )}$:
\begin{equation}
 y_j^{\left( u \right)} = \mathop {{\mathop{\rm argmin}\nolimits} }\limits_{y_j^{\left( u \right)} \in {Y_u}} {\left\| {\varphi \left( {x_i^{\left( u \right)}} \right) - \psi \left( {y_j^{\left( u \right)}} \right)} \right\|^2}
\end{equation}

\section{Experiments}
\subsection{Datasets and setting}
\subsubsection{Datasets.}
To extensive evaluate our method, we adopt four benchmark datasets in this work. The statistics of these datasets are shown in Table \ref{tab:dataset}. Animals with Attributes (AwA1) \cite{Lampert2009Learning} is a coarse-grained dataset that contains 30,475 images from 50 classes of animals. The semantic representation of each class is give as an 85-dimension manually marked attribute vector. In the original AwA dataset  \cite{Lampert2009Learning} (AwA1), the images are not publicly available. In \cite{Xian2017Zero} a new Animals with Attributes2 (AwA2) was introduced with raw images. The AwA2 uses the same 50 animal classes and 85-dimensional attribute vectors as AwA1 data. Both the AwA1 and AwA2 are used to evaluate our model. Caltech UCSD Birds 200-2011 (CUB) \cite{WelinderEtal2010} is a fine-grained dataset that contains 11,788 images from 200 different types of birds annotated with 312 attributes. The original split for zero-shot learning given by \cite{Akata2013Label} including 150 classes for training and 50 classes for testing. SUN Scene Recognition (SUN) \cite{Xiao2010SUN} is a fine-grained dataset that contains 14,340 images from 717 type of scenes annotated with 102 attributes. In the original split \cite{Lampert2014Attribute}, 645 classes are used for training and 72 classes for testing.

In the original split of those datasets, some of the testing categories are subset of the ImageNet \cite{Russakovsky2015ImageNet} categories. When the image features are extracted from ImageNet trained models, it will break the assumption of zero-shot learning that the testing categories are never seen in the training stage. To alleviate this problem, new splits that none of the testing categories coincide with ImageNet categories are proposed in \cite{Xian2017Zero}. The proposed new split method follows the original class number for training and testing. To eliminate confusions and give a fair comparison, this work strictly use the new split datasets, visual features and attributes provided by \cite{Xian2017Zero}. Specifically, the visual features are 2048-dimensional ResNet-101 features and semantic vectors are built by  class-level attributes.

% Caltech UCSD Birds 200-2011 (CUB) \cite{WelinderEtal2010} is a fine-grained dataset that contains 11,788 images from 200 different types of birds annotated with 312 attributes. The original split for zero-shot learning given by \cite{Akata2013Label} including 150 classes for training and 50 classes for testing. 

% SUN Scene Recognition (SUN) \cite{Xiao2010SUN} is a fine-grained dataset that contains 14,340 images from 717 type of scenes annotated with 102 attributes. In the original split \cite{Lampert2014Attribute}, 645 classes are used for training and 72 classes for testing.

% In the original split of those datasets, some of the testing categories are subset of the ImageNet \cite{Russakovsky2015ImageNet} categories. When the image features are extracted from ImageNet trained models, it will break the assumption of zero-shot learning that the testing categories are never seen in the training stage. To alleviate this problem, new splits that none of the testing categories coincide with ImageNet categories are proposed in \cite{Xian2017Zero}. The proposed new split method follows the original class number for training and testing. To eliminate confusions, this work strictly use the new split datasets, visual features and attributes provided by \cite{Xian2017Zero}. Specifically, the visual features are 2048-dimensional ResNet-101 features and attributes are class-level attributes.

\begin{table}
\centering
  \setlength{\tabcolsep}{2.8mm}
  \caption{Statistics of the four zero-shot learning datasets. ``Dims'' is the dimensions of semantic vectors.}
%   \label{tab:freq}
  \begin{tabular}{lcccc}
    \toprule
    Dataset&Dims&Images&Seen Class&Unseen Class\\
    \midrule
    AwA1 \cite{Lampert2009Learning} & 85 &	30475 &	40 & 10\\
    AwA2 \cite{Xian2017Zero} &85&	37322&	40&	10\\
    CUB \cite{WelinderEtal2010}&312&	11788&	150	&50\\
    SUN  \cite{Xiao2010SUN}&102&	14340&	645	&72\\
  \bottomrule
\end{tabular}
\label{tab:dataset}
\end{table}

\subsubsection{Protocols.}
Top-1 accuracy is adopted for the evaluation of single label image classification accuracy. According to the protocol given by Xian et. al \cite{Xian2017Zero}, the zero-shot performance is evaluated based on per-class classification accuracy. Compared with the per-image classification accuracy, this protocol accounts for the imbalances in the target classes and provides a better measurement of the model performance. The evaluation algorithm is as follows:
\begin{equation}
    AC{C_\mathcal { C}} = \frac{1}{{\| \mathcal { C} \|}}\sum\nolimits_{c \in \mathcal { C}} {\frac{{{\rm{\#\  correct\ predictions\ in\ }}c}}{{{\rm{\#\  samples\ in\ }}c}}} 
\end{equation}

As mentioned in the related works, GZSL is a more practical and challenging task, since the search space not only includes unseen classes but also includes seen classes during the evaluation stage. To evaluate the performance on GZSL, we use the harmonic mean of seen and unseen accuracy as existing works:
\begin{equation}
    H{\rm{ = }}\frac{{{\rm{2}}AC{C_{tr}} \times AC{C_{ts}}}}{{AC{C_{tr}} + AC{C_{ts}}}}
\end{equation}
where $AC{C_{tr}}$ and $AC{C_{ts}}$ represent the recognition accuracy of images from seen and unseen classes respectively. The harmonic mean pays more attention on the overall recognition performance, i.e. both of the unseen recognition and the seen recognition, and is able to avoid the effect of much higher seen class accuracy.

\subsubsection{Implementation details.}
The proposed framework is implemented using PyTorch \footnote{https://pytorch.org/}. For the visual prototype based method, we initialize a visual prototype with the average vectors of visual features for each class. Then the visual prototypes can be updated with the loss function Eq. (\ref{eq:loss_proto}). With learned visual prototypes, the embedding framework for the semantic vectors can be learned  with the loss function Eq. (\ref{eq:loss_proto_emb}). In practice, we train the visual prototypes and embedding framework in an alternate way. Specifically, every 500 iterations for prototypes learning followed by 1000 iterations for embedding framework training. Details of the training parameters are shown in table \ref{tab:setting_proto}.

\begin{table}[]
	\centering
	\setlength{\tabcolsep}{1.5mm}
	\caption{Training sets for different datasets in prototype based method. LR denotes learning rate and HL denotes the dimension of the hidden layer.}
	\begin{tabular}{lcccc}
	\toprule
		\multirow{2}{*}{Datasets} &\multirow{2}{*}{Batch size}  & {Prototypes learning} & \multicolumn{2}{c}{Embedding framework}          \\ 
		\cline{3-5} 
		~&~& LR     &LR      & \multicolumn{1}{c}{HL} \\
		\midrule
		\multicolumn{1}{l}{AwA1} &100&	1e-5&	1e-6&	800           \\ 
		AwA2&	100&	1e-5&	1e-6&	800  \\
		CUB	&100&	1e-5&	1e-4&	1000   \\
		SUN	&100&	1e-5&	1e-6&	2048   \\
		\bottomrule
	\end{tabular}
	\label{tab:setting_proto}
\end{table}

In the visual data structure optimization based method, both of attributes and visual features are transformed into the intermediate embedding space with a two-layer multilayer perceptron. Since the hubness problem \cite{Angeliki2014Hubness} can be suppressed effectively when the visual space works as the embedding space \cite{Zhang2017Learning}, we expect the intermediate embedding space to be closer with the visual space. To this end, the weights ${V_1}$ and ${V_2}$ are initialized to be unit diagonal matrices, so that the initial visual embedding features will be the same as the original visual features. The learning rate for ${V_1}$ and ${V_2}$ is smaller than that for ${W_1}$ and ${W_2}$. With these settings, we can make sure that the learned intermediate embedding space is closer with the visual space. The learning rates and other parameters for training the model on different datasets are listed in table \ref{tab:setting_stur}.

\begin{table}[]
	\centering
	\setlength{\tabcolsep}{3mm}
	\caption{Training sets for different datasets in visual data optimization based method. LR denotes learning rate and HL denotes the dimension of the hidden layer.}
	\begin{tabular}{lccccc}
	\toprule
		\multirow{2}{*}{Datasets} &\multirow{2}{*}{Batch size}  & \multicolumn{2}{c}{Semantic branch} & \multicolumn{2}{c}{Visual branch}          \\ 
		\cline{3-6} 
		~&~& LR        & HL  &LR      & \multicolumn{1}{l}{HL} \\
		\midrule
		\multicolumn{1}{l}{AwA1} &100&	1e-5&	800&	1e-7&	2048           \\ 
		AwA2&	100&	1e-5&	800&	1e-7&	2048  \\
		CUB	&64&	1e-4&	1000&	1e-6&	2048   \\
		SUN	&100&	1e-4&	1600&	1e-6&	2048   \\
		\bottomrule
	\end{tabular}
	\label{tab:setting_stur}
\end{table}

\subsection{Compare with the state-of-the-art}
A wide range of existing ZSL models are selected for the performance comparison. Among these models, DAP and IAP \cite{Lampert2014Attribute}, CONSE \cite{Norouzi2013Zero}, CMT \cite{Socher2013Zero}, SSE \cite{Zhang2016Zero}, LATEM \cite{Xian2016Latent}, ALE \cite{Akata2016Label}, DEVISE \cite{Frome2013Devise}, SJE \cite{Akata2015Evaluation}, ESZSL \cite{Romera-Paredes2015An}, SYNC \cite{Changpinyo2016Synthesized}, SAE \cite{Kodirov2017Semantic} and GFZSL \cite{Verma2018Verma} are mostly compared in lots of recent works, and the corresponding results are directly taken from \cite{Xian2017Zero}. Note that even though those methods originally adopted different visual deep features or evaluation methods, they were re-evaluated by \cite{Xian2017Zero} using the unified features and evaluation protocol. For a fair comparison, we also exactly utilize the same features and evaluation protocol in this work. Besides, some recent works, including TVN \cite{Zhang2018Triple}, LESAE \cite{Liu2018Zero}, PSR \cite{Annadani2018Preserving}, DCN \cite{Liu2018Generalized}, and MLSE \cite{ding2019marginalized} are also considered. These recent works have excellent performances, but no one can beat all the others on all four datasets. For instance, in the task of ZSL, the MLSE \cite{ding2019marginalized} achieves the highest accuracy on datasets of CUB and SUN, whereas the best performances on the datasets of AwA1 and AwA2 are achieved by TVN \cite{Zhang2018Triple} and LESAE \cite{Liu2018Zero} respectively. In this work, we compare the proposed methods with the above on the both tasks of ZSL and GZSL. Since this work adopts the original global visual features and no data augmentation strategy is used, those works that focus on data augmentation \cite{xian2019f,schonfeld2019generalized,huang2019generative,bulent2019gradient,li2019leveraging} or try to get more distinctive visual features \cite{xie2019attentive} are not considered in the comparison.

\subsubsection{Performance on zero-shot learning}
Table \ref{tab:comparsion} presents the zero-shot learning Top-1 accuracy on the four datasets. The mehtods ``Proposed-SRS'' and ``Proposed-BRS'' represent the methods with visual data structure optimization, of which the former is the the proposed method using the simple ranking loss and visual structure optimizing loss formulated by Eq. (\ref{eq:Lsrs}), and the latter indicates the method formulated by Eq. (\ref{eq1:Lbrs}) which adopts the bi-directional ranking loss along with the visual structure optimizing loss. The ``Proposed-VPB'' represents the visual prototype based method. 

Compared with previous methods, the methods with visual structure optimization show outstanding performance on the ZSL task. As shown in Table \ref{tab:comparsion}, the proposed SRS and BRS outperform all existing methods except on the CUB dataset. Specifically, SRS and BRS exceed the best competitive method TVN \cite{Zhang2018Triple} by 1.2$\%$ and 1.4$\%$ respectively on the dataset of AwA1. On the AwA2 dataset, SRS and BRS outperform PSR \cite{Annadani2018Preserving} significantly, and the gains are more than 5$\%$ in Top-1 accuracy. Compared with the best existing method LESAE \cite{Liu2018Zero}, we still enjoy at least 1$\%$ gains. On the dataset of SUN, the proposed methods achieve the best accuracy gotten by MLSE \cite{ding2019marginalized}. On the dataset of CUB, although the proposed visual structure optimization based methods are inferior to several works \cite{Zhang2018Triple,Annadani2018Preserving,Liu2018Generalized,ding2019marginalized}, they are superior to other methods. Compared with the simple ranking loss, the bi-directional ranking loss shows only a slight superiority on ZSL. Specifically, the largest gap between the Proposed-SRS and Proposed-BRS is only 0.7$\%$ appearing on the CUB dataset. This phenomenon will be discussed in the section of Discussion. 

The proposed method based on the visual prototype achieves more outstanding performance. As presented in Table \ref{tab:comparsion}, the proposed VPB method only get lower accuracy than SRS and BRS on the CUB dataset, and the same on the dataset of SUN. On datasets AwA1 and AwA2, the VPB exceed the BRS by 3.1$\%$ and 3.8$\%$. Compared with the previous best accuracy on AwA1 achieved by TVN \cite{Zhang2018Triple}, the VPB get 3.5$\%$ gains, and the raise is 5.4$\%$ on the dataset AwA2 compared with the previous best method  LESAE \cite{Liu2018Zero}.

\begin{table}
\centering
  \caption{Zero-shot learning results on AwA1, AwA2, CUB and SUN. The results are measured in Top-1 accuracy (\%).}
%   \label{tab:freq}
  \setlength{\tabcolsep}{4mm}
 
  \begin{tabular}{lcccc}
    \toprule
    Methods&AwA1 &	AwA2 &	CUB	 &SUN\\
    \midrule
DAP \cite{Lampert2014Attribute} & 44.1 & 46.1 &	40.0 &	39.9\\
IAP \cite{Lampert2014Attribute} & 35.9 & 35.9 &	24.0 &	19.4\\
CONSE \cite{Norouzi2013Zero} & 45.6 &	44.5 &	34.3 &	38.8\\
CMT \cite{Socher2013Zero} & 39.5 &	37.9 &	34.6 &	39.9\\
SSE \cite{Zhang2016Zero} & 60.1 &	61.0 &	43.9 &	51.5\\
LATEM \cite{Xian2016Latent}& 55.1 &	55.8 &	49.3 &	55.3\\
ALE \cite{Akata2016Label}& 59.9 &	62.5 &	54.9 &	58.1\\
DEVISE \cite{Frome2013Devise}& 54.2 &	59.7 &	52.0 &	56.5\\
SJE \cite{Akata2015Evaluation}& 65.6 &	61.9 &	53.9 &	53.7\\
ESZSL \cite{Romera-Paredes2015An}& 58.2 &	58.6 &	53.9 &	54.5\\
SYNC \cite{Changpinyo2016Synthesized}& 54.0 &	46.6 &	55.6 &	56.3\\
SAE \cite{Kodirov2017Semantic}& 53.0 &	54.1 &	33.3 &	40.3\\
GFZSL \cite{Verma2018Verma}& 68.3 &	63.8 &	49.3 &	60.6\\
TVN \cite{Zhang2018Triple}& 68.8	&---	& 58.1  &	60.7  \\
LESAE \cite{Liu2018Zero}& 66.1&	68.4&	53.9&	60.0\\
PSR \cite{Annadani2018Preserving}& ---&	63.8&	56.0&	61.4\\
DCN \cite{Liu2018Generalized} & 65.2&---&56.2&61.8\\
MLSE \cite{ding2019marginalized}& --- & 67.8 & $\mathbf{64.2}$ & $\mathbf{62.8}$\\
\midrule	
Proposed-SRS&	70.0&	69.4&	55.0&	$\mathbf{62.8}$\\
Proposed-BRS&	70.2&70.0&	55.7&	$\mathbf{62.8}$\\
Proposed-VPB&	$\mathbf{72.3}$&	$\mathbf{73.8}$&	52.1&	$\mathbf{62.8}$\\
  \bottomrule
\end{tabular}
\label{tab:comparsion}
\end{table}

% the former best result achieved by LESAE \cite{Liu2018Zero} is 68.4$\%$ which is really outstanding on this dataset, as it is 4.2$\%$ higher than the result in the second place got by RN \cite{Sung2018Learning}, while both our proposed methods outperform the LESAE \cite{Liu2018Zero} on this dataset and exceed it 1.0$\%$ and 1.6$\%$ by Proposed-SRS and Proposed-BRS respectively. Both of the proposed methods get 1.4$\%$ higher than the former best result on the SUN datasets which is 61.4$\%$ achieved by PSR \cite{Annadani2018Preserving}. 

%在结果比较部分去掉了本文两种方法的讨论，该部分在discussion部分
% Excepting get the same performance on the SUN dataset, the proposed method based on the bi-directional ranking loss outperforms the simple ranking loss based method on all datasets. However, the advantage using the bi-directional ranking loss is slight, and the largest gap between the Proposed-SRS and Proposed-BRS is only 0.7 percent on the CUB dataset. One probable reason is that the semantic vectors natively have a good data structure, and when the data structure is optimized on the visual features, the simple ranking loss is enough to get outstanding performance.
\subsubsection{Performance on generalized zero-shot learning}
Table \ref{tab:GZSL} reports the result of generalized zero-shot learning on the four datasets. ts refers $ACC_{ts}$ in which the testing samples belong to unseen classes, and tr refers $ACC_{tr}$ wherein the testing samples belongs to seen classes. The target labels for the evaluation of both ts and tr are all classes including seen and unseen. High accuracy on tr and low accuracy on ts means that the method performs well on the seen classes but fails to work well on the unseen classes, which implies the overfitting on the seen classes. The harmonic mean (H) of tr and ts gives the comprehensive evaluation on the GZSL task.

As shown in Table \ref{tab:GZSL}, the methods based on visual data structure optimization seem not get obvious superiority on the GZSL task compared with previous state-of-the-art methods. Nevertheless, they still show comparable results with most recent works. Practically, the BRS achieves the best harmonic mean accuracy 38.3\% on AwA2 compared with all recently proposed methods. On datasets AwA1 and CUB, the proposed BRS is only inferior to DCN \cite{Liu2018Generalized}  with small gaps. The BRS is better than SRS on all the four datasets. 

The stimulating is that the proposed method based on the visual prototypes VPB achieves considerable improvement compared with all recently proposed methods. Specifically, the proposed VPB gives a harmonic mean accuracy of 55.6\% on AwA1 which is the best result among all the reported methods, and it higher than the previous best result by 16.5\%, which is indeed a huge improvement. The huge increment also appears on the AwA2, where the harmonic mean accuracy is 53.8\%,  16.8\% better than the exiting best method. One datasets CUB and SUN, the proposed VPB also shows inspiring performance, and it obtains the best accuracy of 40.7\% and 37.3\% on these two datasets respectively, which are 2\% and 7.1\% better than the next best methods on CUB and SUN.

\begin{table*}[]
\centering
\caption{generalized zero-shot learning results on AwA1, AwA2, CUB and SUN. The results are measured in top-1 accuracy (\%).}
\setlength{\tabcolsep}{3.5mm}
\begin{tabular}{lccc|ccc|ccc|ccc}
\toprule
       & \multicolumn{3}{c|}{AwA1} & \multicolumn{3}{c|}{AwA2} & \multicolumn{3}{c|}{CUB} & \multicolumn{3}{c}{SUN} \\ \midrule
Method & ts      & tr     & H      & ts      & tr     & H      & ts     & tr     & H      & ts     & tr     & H     \\ \midrule
DAP \cite{Lampert2014Attribute}        & 0.0     & $\mathbf{88.7}$   & 0.0    & 0.0     & 84.7   & 0.0    & 1.7    & 67.9   & 3.3    & 4.2    & 25.1   & 7.2   \\
IAP \cite{Lampert2014Attribute}        & 2.1     & 78.2   & 4.1    & 0.9     & 87.6   & 1.8    & 0.2    & $\mathbf{72.8}$   & 0.4    & 1.0    & 37.8   & 1.8   \\
CONSE \cite{Norouzi2013Zero}           & 0.4     & 88.6   & 0.8    & 0.5     & $\mathbf{90.6}$   & 1.0    & 1.6    & 72.2   & 3.1    & 6.8    & 39.9   & 11.6  \\
CMT \cite{Socher2013Zero}              & 0.9     & 87.6   & 1.8    & 0.5     & 90.0   & 1.0    & 7.2    & 49.8   & 12.6   & 8.1    & 21.8   & 11.8  \\
SSE \cite{Zhang2016Zero}               & 7.0     & 80.5   & 12.9   & 8.1     & 82.5   & 14.8   & 8.5    & 46.9   & 14.4   & 2.1    & 36.4   & 4.0   \\
LATEM \cite{Xian2016Latent}            & 7.3     & 71.7   & 13.3   & 11.5    & 77.3   & 20.0   & 15.2   & 57.3   & 24.0   & 14.7   & 28.8   & 19.5  \\
ALE \cite{Akata2016Label}              & 16.8    & 76.1   & 27.5   & 14.0    & 81.8   & 23.9   & 23.7   & 62.8   & 34.4   & 21.8   & 33.1   & 26.3  \\
DEVISE \cite{Frome2013Devise}          & 13.4    & 68.7   & 22.4   & 17.1    & 74.7   & 27.8   & 23.8   & 53.0   & 32.8   & 16.9   & 27.4   & 20.9  \\
SJE \cite{Akata2015Evaluation}         & 11.3    & 74.6   & 19.6   & 8.0     & 73.9   & 14.4   & 23.5   & 59.2   & 33.6   & 14.7   & 30.5   & 19.8  \\
ESZSL \cite{Romera-Paredes2015An}      & 6.6     & 75.6   & 12.1   & 5.9     & 77.8   & 11.0   & 12.6   & 63.8   & 21.0   & 11.0   & 27.9   & 15.8  \\
SYNC \cite{Changpinyo2016Synthesized}  & 8.9     & 87.3   & 16.2   & 10.0    & 90.5   & 18.0   & 11.5   & 70.9   & 19.8   & 7.9    & 43.3   & 13.4  \\
SAE \cite{Kodirov2017Semantic}         & 1.8     & 77.1   & 3.5    & 1.1     & 82.2   & 2.2    & 7.8    & 54.0   & 13.6   & 8.8    & 18.0   & 11.8  \\
GFZSL \cite{Verma2018Verma}            & 1.8     & 80.3   & 3.5    & 2.5     & 80.1   & 4.8    & 0.0    & 45.7   & 0.0    & 0.0    & 39.6   & 0.0   \\
TVN \cite{Zhang2018Triple}             & 27.0    & 67.9   & 38.6   & ---     & ---    & ---    & 26.5   & 62.3   & 37.2   & 22.2   & 38.3   & 28.1  \\
LESAE \cite{Liu2018Zero}               & 19.1    & 70.2   & 30.0   & 21.8    & 70.6   & 33.3   & 24.3   & 53.0   & 33.3   & 21.9   & 34.7   & 26.9  \\
PSR \cite{Annadani2018Preserving}      &---      & ---    & ---    & 20.7    & 73.8   & 32.3   & 24.6   & 54.3   & 33.9   & 20.8   & 37.2   & 26.7  \\
DCN \cite{Liu2018Generalized}          & 25.5    & 84.2   & 39.1   & ---     & ---    &  ---   & 28.4   & 60.7   & 38.7   & 25.5   & 37.0   & 30.2  \\
MLSE \cite{ding2019marginalized}       &---      & ---    & ---    &23.8     &83.2    & 37.0   & 22.3   & 71.6   & 34.0    & 20.7   & 36.4  &26.4  \\
\midrule
Proposed-SRS                           & 22.7    & 82.3   & 35.6   & 22.8    & 83.2   & 35.8   & 24.2   & 59.1   & 34.3   & 18.6   & 39.4   & 25.3  \\
Proposed-BRS                           & 25.5    & 83.0   & 39.0   & 24.8    & 84.5   & 38.3   & 27.7   & 58.6   & 37.6   & 19.8   & 35.7   & 25.5  \\
Proposed-VPB                           & $\mathbf{43.2}$    & 77.8   & $\mathbf{55.6}$   & $\mathbf{40.4}$    & 80.8   & $\mathbf{53.8}$   & $\mathbf{30.7}$   & 60.5   & $\mathbf{40.7}$   & $\mathbf{28.5}$   & $\mathbf{53.9}$   & $\mathbf{37.3}$  \\
\bottomrule
\label{tab:GZSL}
\end{tabular}
\end{table*}

\subsection{Ablation study}
In this section, we present ablation analysis of the proposed methods on the four datasets. For the proposed methods SRS and BRS, the effectiveness of optimizing the visual structure are conducted. In the visual prototypes based method, the superiority of the learned visual prototypes are analyzed.

\subsubsection{Effectiveness of optimizing the visual structure}
To demonstrate the effectiveness of the visual structure optimizing loss, Table \ref{tab:abl_structure} displays the performance of the proposed models and corresponding algorithms without it on the task of ZSL. SR and SRS denote the simple ranking loss without and with visual structure optimizing loss, respectively. BR and BRS indicate the bi-directional ranking loss without and with visual structure optimizing loss, respectively.

As shown in this table, the item of the visual structure optimizing loss plays a very  important role in the proposed framework. The proposed methods are obviously superior to the corresponding methods without optimizing the visual data structure. Specifically, the gap between SR and SRS reaches 7.9$\%$ on the CUB dataset, and their minimum gap  is 1.1$\%$ on the AwA2 dataset. Similarly, the largest gap between BR and BRS achieves 7.7 $\%$ on the CUB dataset. The minimum gap appears on the AwA1 dataset, where the BRS is 2.5$\%$ higher than the BR. Compared with the results on AwA1 and AwA2 datasets, the visual structure optimization shows a more obvious advantage on the CUB and SUN datasets. According to Table \ref{tab:dataset}, these two datasets have more classes compared with the other two datasets. Since the more classes the more chaotic of the visual data structure will be, the proposed method can effectively alleviate this problem.

\begin{table}[h]
\centering
\caption{Effectiveness of visual structure optimizing on ZSL. The results are measured in top-1 accuracy (\%).}
\setlength{\tabcolsep}{5mm}
\begin{tabular}{lcc|cc}
\toprule
     & SR   & SRS  & BR   & BRS  \\ \midrule
AwA1 & 68.0  & 70.0  & 67.7 & 70.2 \\
AwA2 & 68.3 & 69.4 & 66.6 & 70.0   \\
CUB  & 47.1 & 55.0   & 48.0   & 55.7 \\
SUN  & 55.3 & 62.8 & 55.6 & 62.8 \\ \bottomrule
\end{tabular}
\label{tab:abl_structure}
\end{table}

\begin{figure}[htb]
\centering
\subfigure[AwA1]{
 \begin{minipage}[t]{0.45\linewidth}
 \centering
 \includegraphics[scale = 0.18]{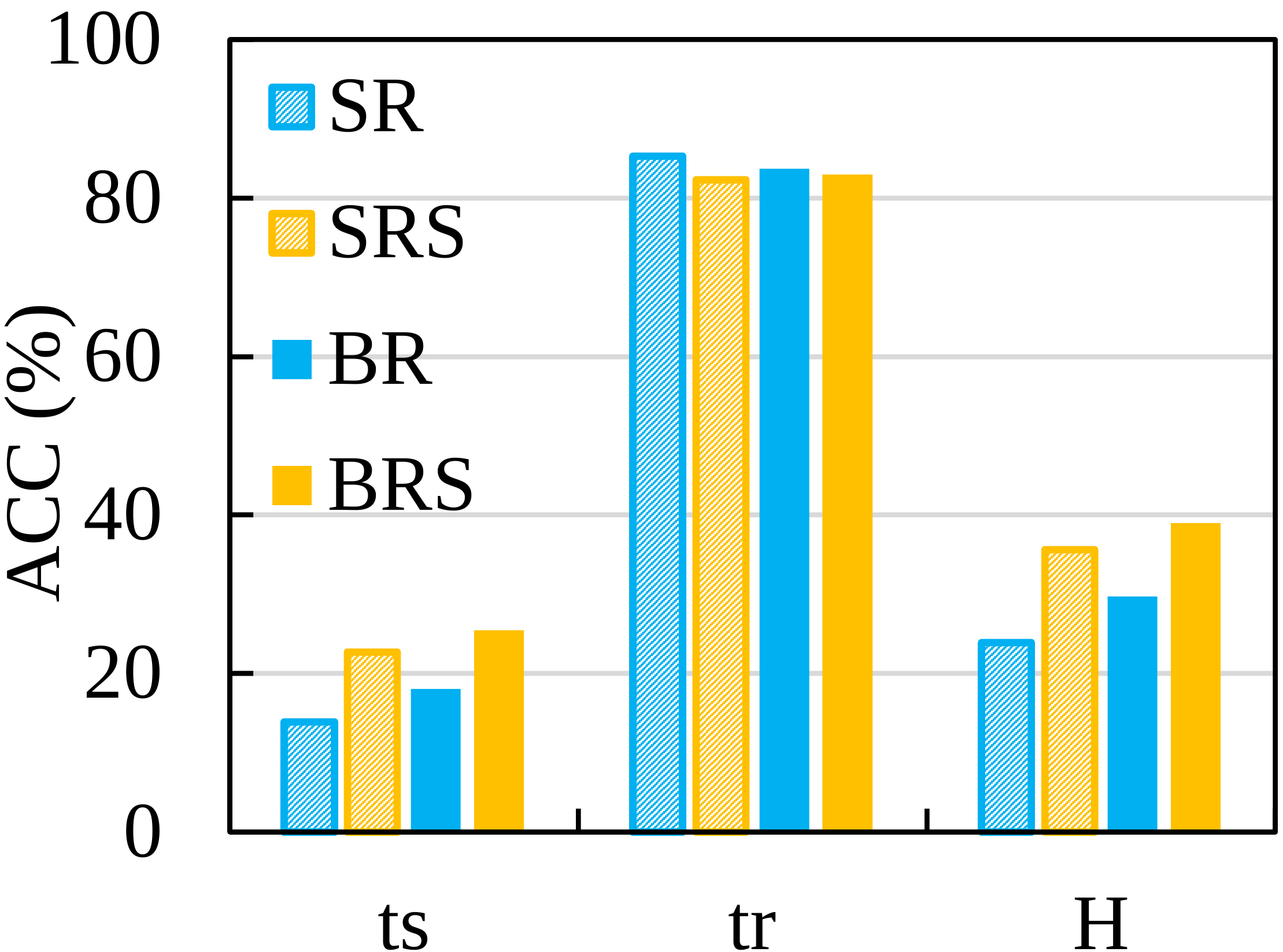}
 \label{fig:abl_structure_GZSL_a}
 \end{minipage}
 }
% \hspace{0.1in}
\subfigure[AwA2]{
  \begin{minipage}[t]{0.45\linewidth}
  \includegraphics[scale = 0.18]{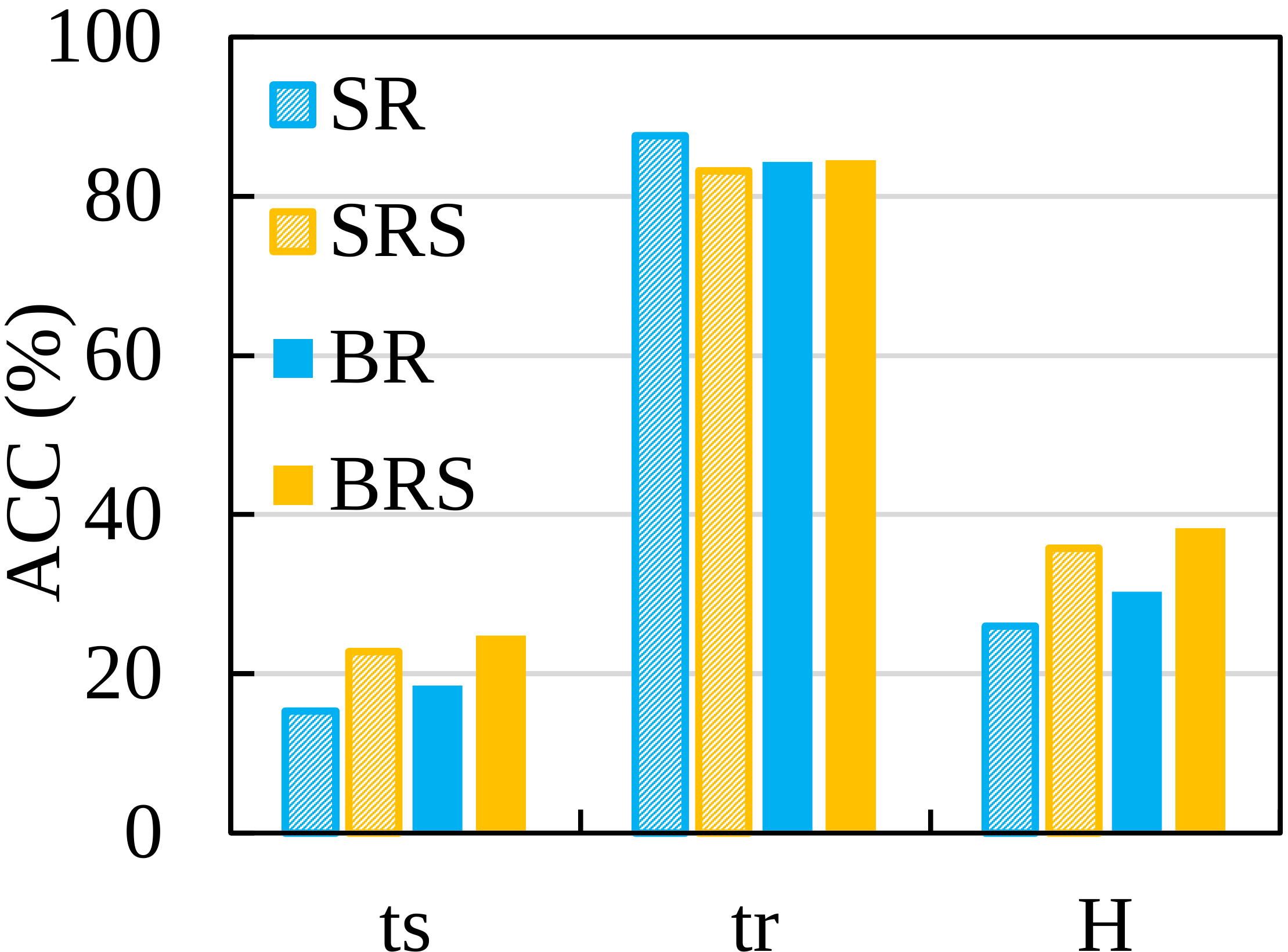}
  \centering
  \label{fig:abl_structure_GZSL_b}
  \end{minipage}
  }\\
  \subfigure[CUB]{
  \begin{minipage}[t]{0.45\linewidth}
  \includegraphics[scale = 0.18]{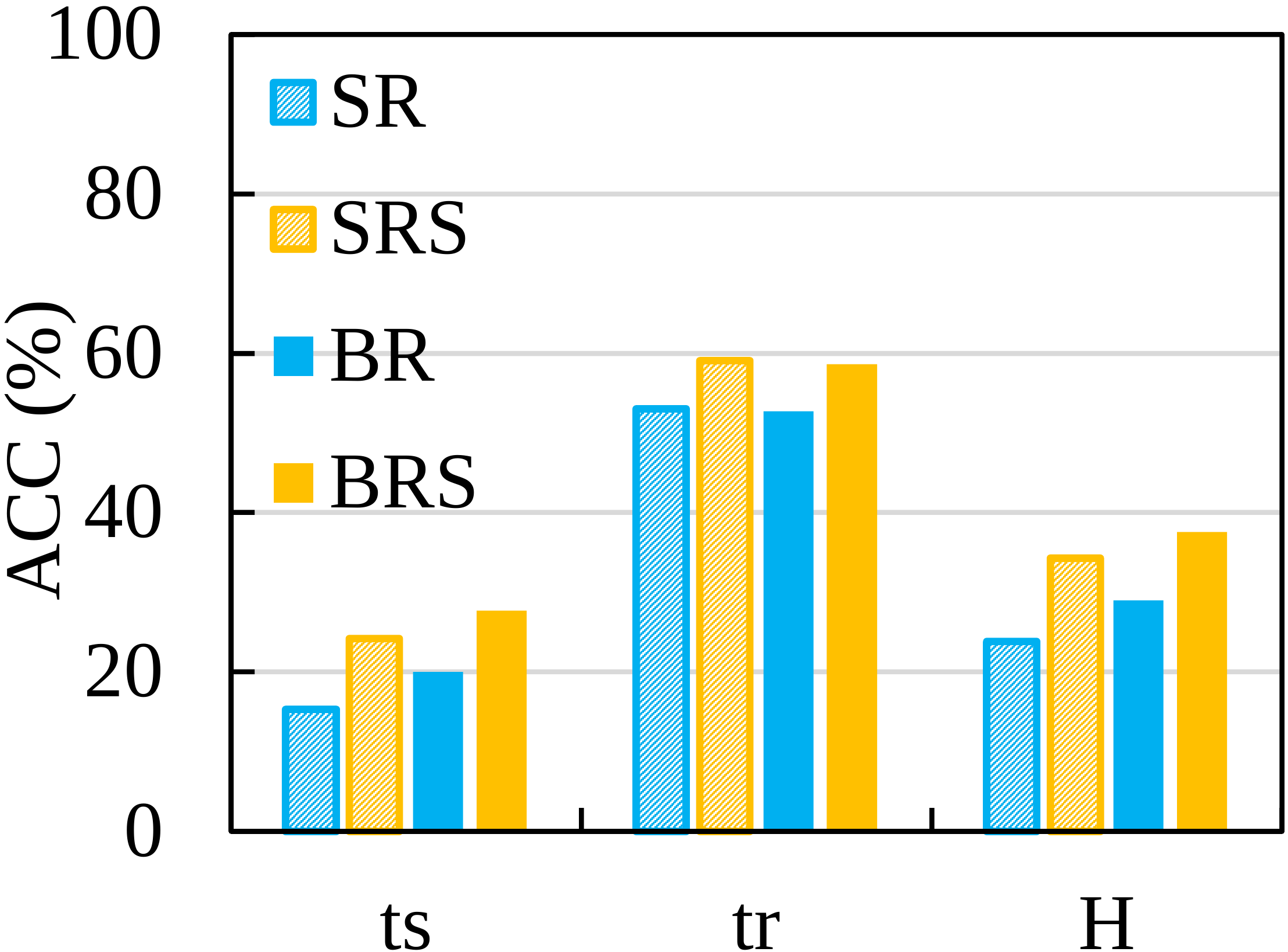}
  \centering
  \label{fig:abl_structure_GZSL_c}
  \end{minipage}
  }
  \subfigure[SUN]{
  \begin{minipage}[t]{0.45\linewidth}
  \includegraphics[scale = 0.18]{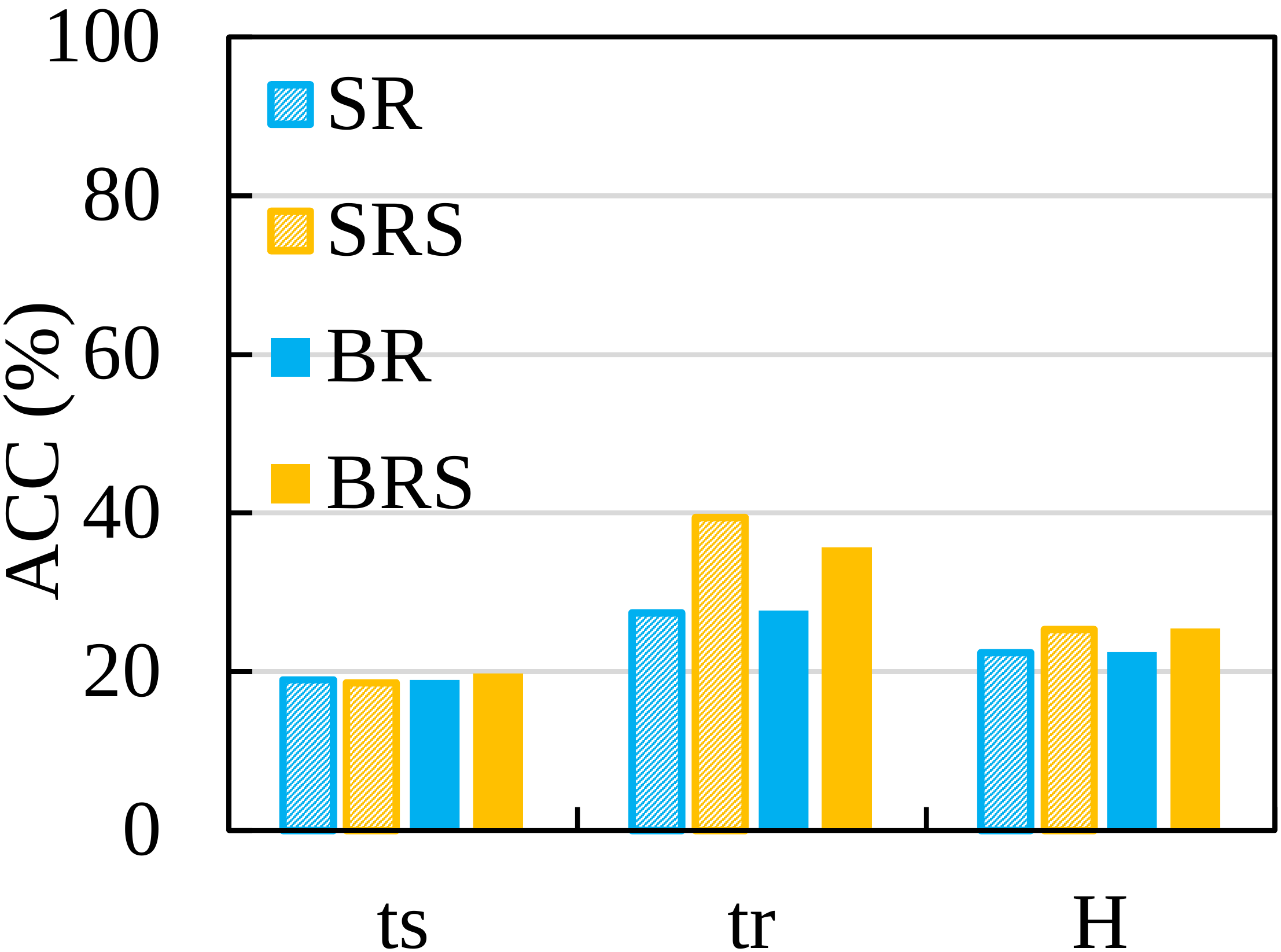}
  \centering
  \label{fig:abl_structure_GZSL_d}
  \end{minipage}
  }
 \centering
 \caption{Effectiveness of the visual structure optimizing function on GZSL. SR and SRS denote the simple ranking loss without and with visual structure optimizing loss, respectively. BR and BRS indicate the bi-directional ranking loss without and with visual structure optimizing loss, respectively.}
 \label{fig:abl_structure_GZSL}
 \end{figure}

The visual structure optimizing loss also plays an important role in the task of GZSL. As illustrated in Fig. \ref{fig:abl_structure_GZSL}, the harmonic mean accuracy of methods with structure optimizing loss consistently outperform the corresponding methods without structure optimizing loss with improvement from 2.9\% to 11.7\%. On the datasets AwA1, AwA2, and CUB, the improvement of H achieved by the visual structure optimization is more than 8.0\%, whereas the minimum increment appears on SUN, which is different with the performance of visual structure optimization on the task of ZSL. The reason is that, in the more realistic task of GZSL, the ability of models to avoid overfitting on the seen classes plays a critical role. The models without visual structure optimizing loss tend to overfit on the datasets with small classes, e.g. AwA1 and AwA2. As shown in Fig. \ref{fig:abl_structure_GZSL_a} and Fig. \ref{fig:abl_structure_GZSL_b}, the SR and BR get higher tr accuracy but very low ts accuracy compared with SRS and BRS.

\subsubsection{Performance of learnable visual prototypes}
Averaging the visual features of one class to get the visual centroid vector is an intuitive way to get the visual prototype. To test the superiority of the learnable visual prototypes proposed in this work compared with the visual centroid vectors, we compare the performance of centroid based and learned prototype based method on ZSL and GZSL.

Table \ref{tab:abl_proto_zsl} presents the comparison of visual centroid based (VCB) and learned prototype based  methods on the task of ZSL. As shown in this table, the learned prototype based method outperforms the centroid based method on most of datasets but fails on SUN. This only failure of learned prototype on SUN is caused by the small proportion of test classes. From Table \ref{tab:dataset}, one can see that the test class number only accounts to 10\%, which means the test visual features tend to be discriminative, since there will be less possible for the intersection of instance features from different classes. In that case, the centroids naturally give better performance. After all, they are centorids of each class. Anyway, the more persuasive performance of learned prototypes will be conducted on the task of GZSL.

\begin{table}[]
\centering
\caption{Comparison of visual centroid based and learned prototype based performance on ZSL. The results are measured in top-1 accuracy (\%).}
\setlength{\tabcolsep}{8mm}
\begin{tabular}{lcc}
\toprule
     & VCB & VPB \\ \midrule
AwA1 & 68.1     & 72.3              \\ 
AwA2 & 67.6     & 73.8              \\ 
CUB  & 51.5     & 52.1              \\ 
SUN  & 63.4     & 62.8              \\ \bottomrule
\end{tabular}
\label{tab:abl_proto_zsl}
\end{table}

\begin{figure}[htb]
\centering
\subfigure[AwA1]{
 \begin{minipage}[t]{0.45\linewidth}
 \centering
 \includegraphics[scale = 0.18]{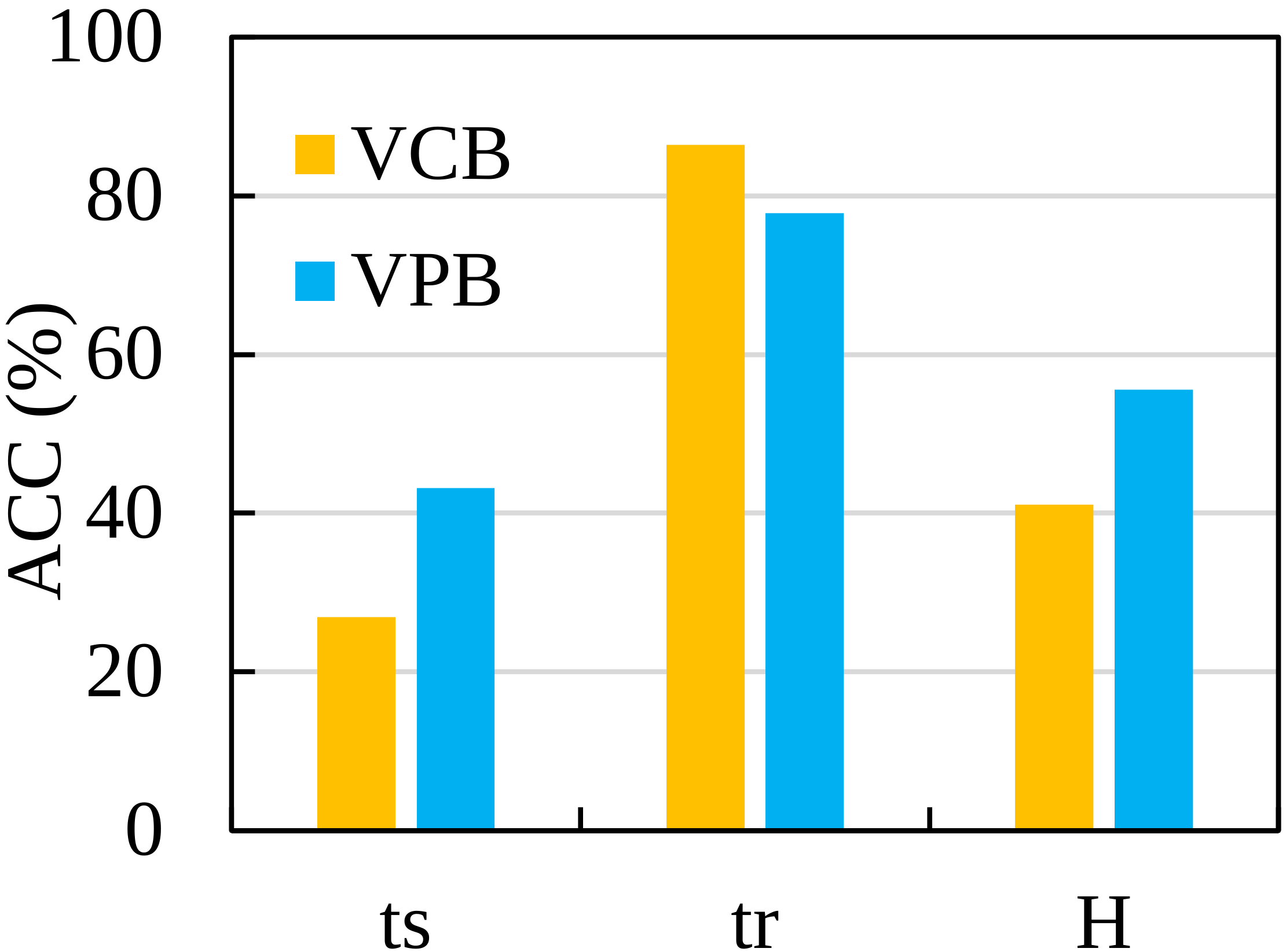}
 \label{fig:abl_proto_GZSL_a}
 \end{minipage}
 }
% \hspace{0.1in}
\subfigure[AwA2]{
  \begin{minipage}[t]{0.45\linewidth}
  \includegraphics[scale = 0.18]{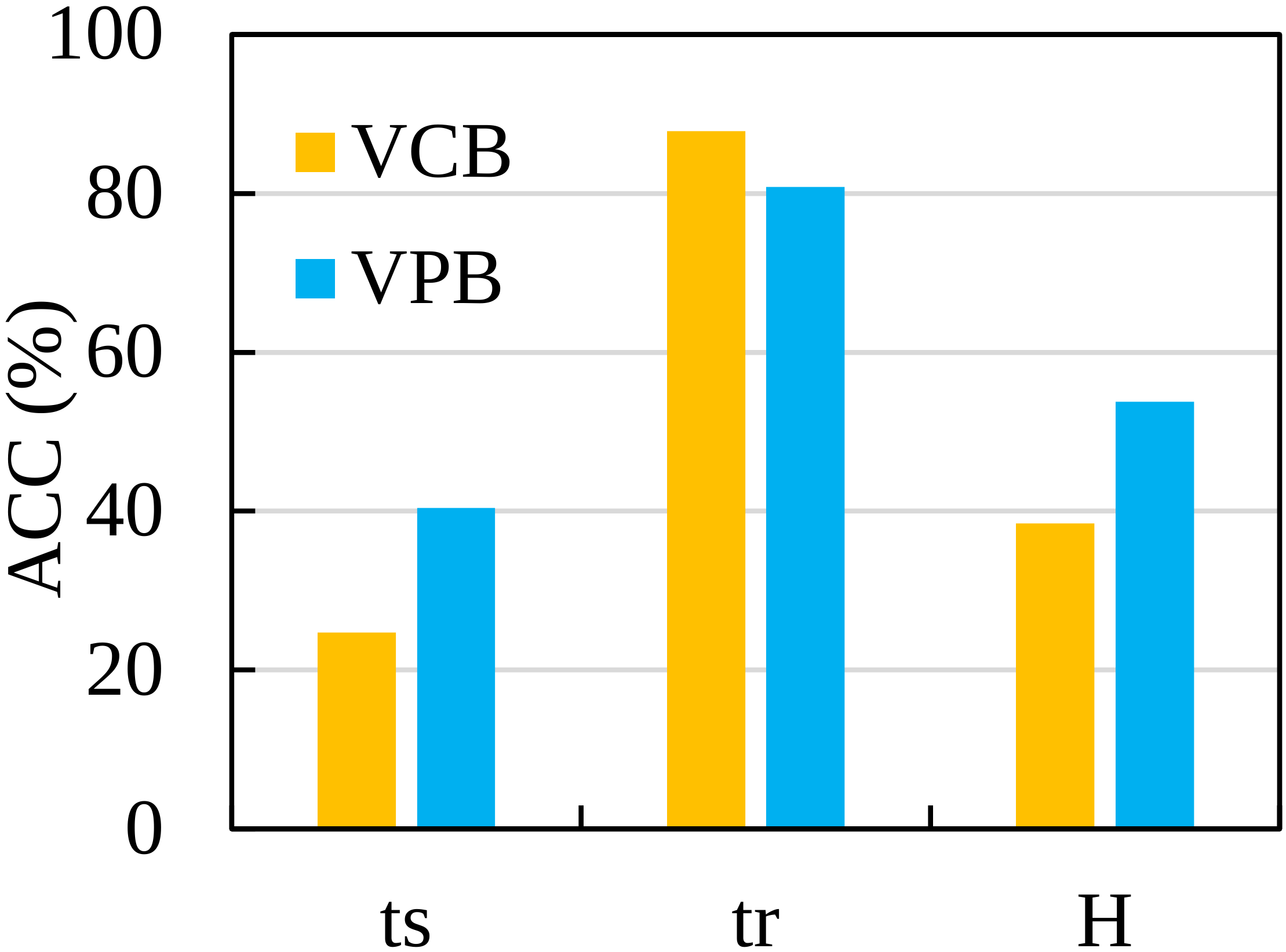}
  \centering
  \label{fig:abl_proto_GZSL_b}
  \end{minipage}
  }\\
  \subfigure[CUB]{
  \begin{minipage}[t]{0.45\linewidth}
  \includegraphics[scale = 0.18]{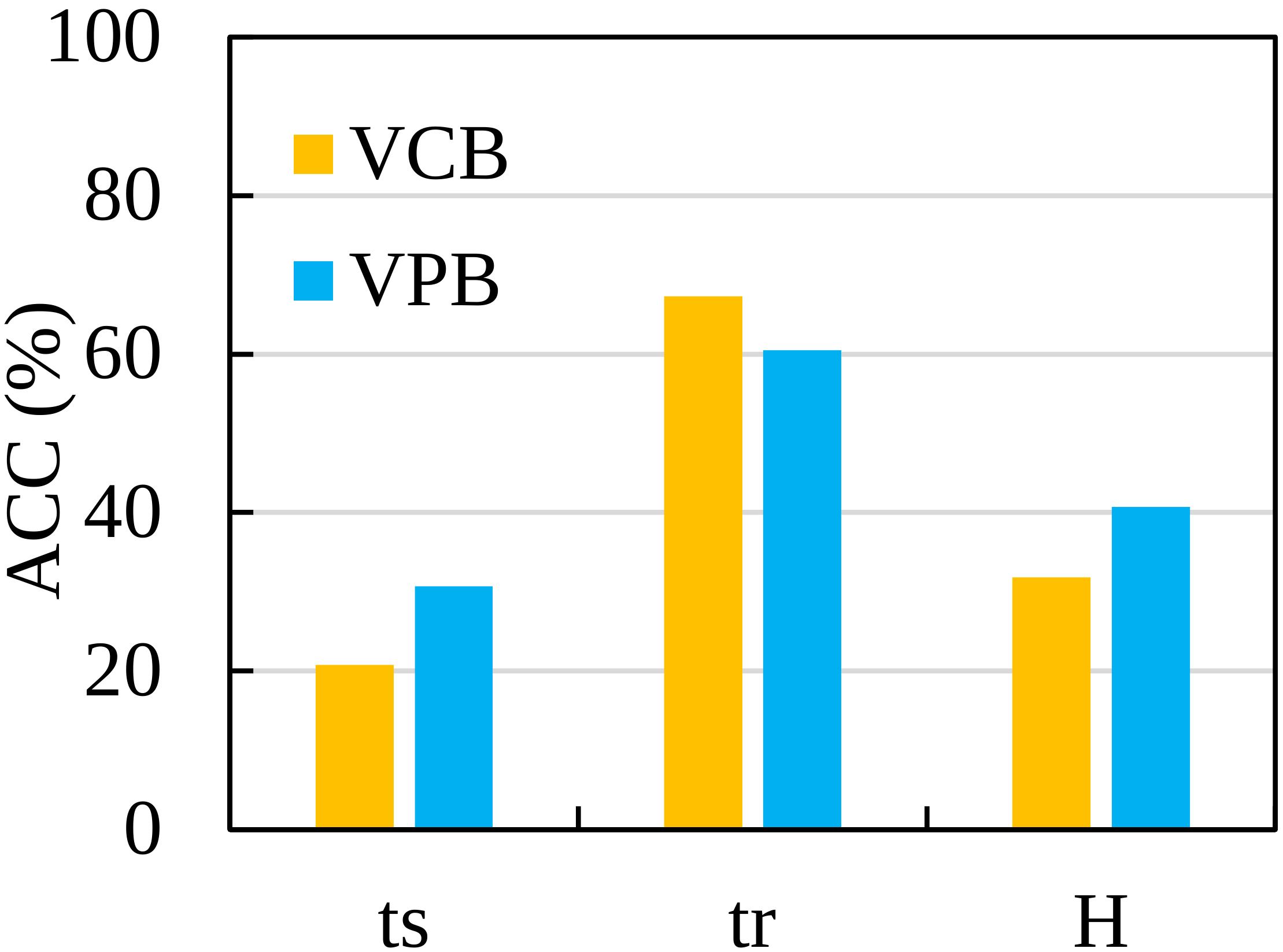}
  \centering
  \label{fig:abl_proto_GZSL_c}
  \end{minipage}
  }
  \subfigure[SUN]{
  \begin{minipage}[t]{0.45\linewidth}
  \includegraphics[scale = 0.18]{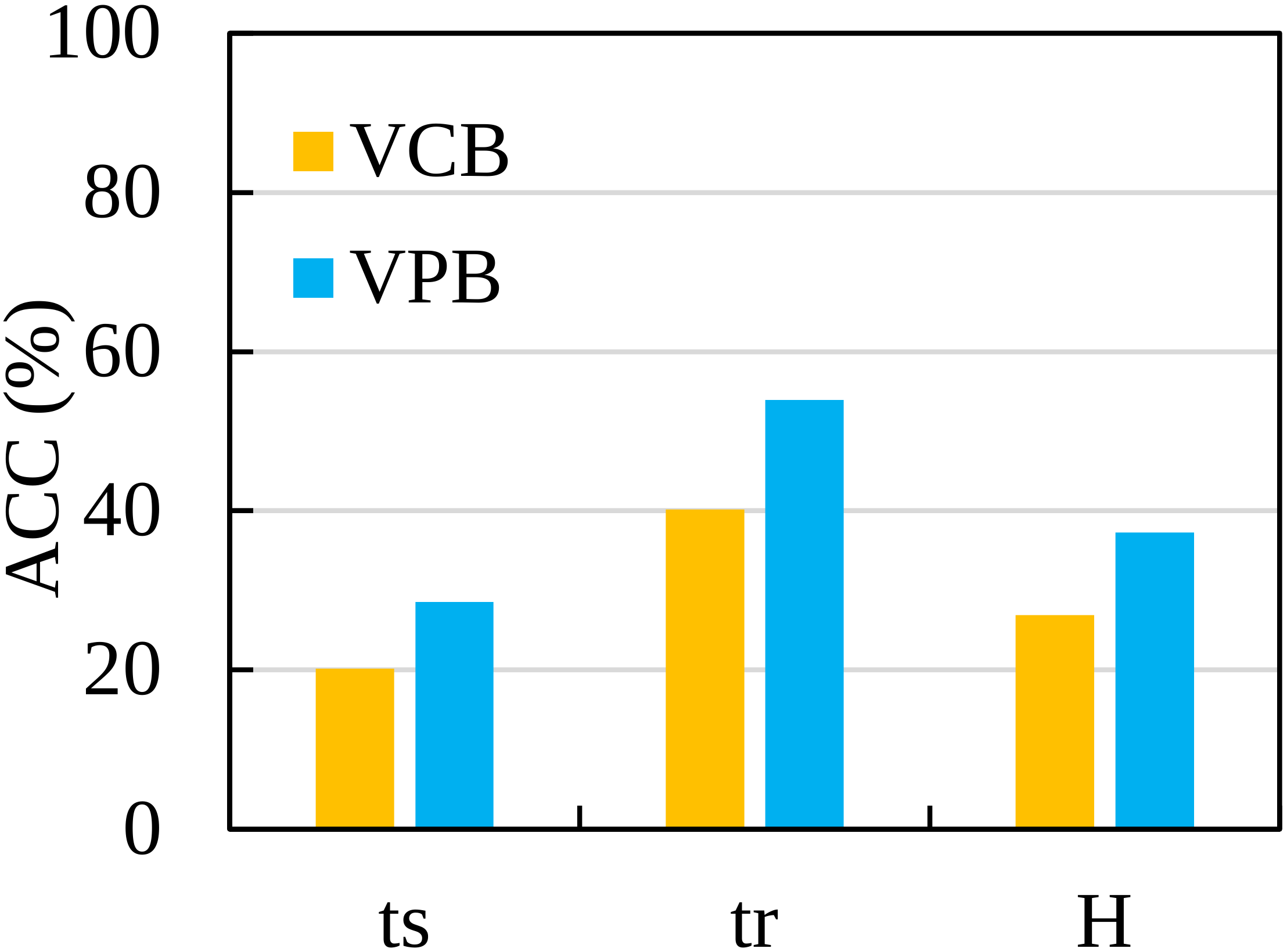}
  \centering
  \label{fig:abl_proto_GZSL_d}
  \end{minipage}
  }
 \centering
 \caption{Comparison of visual centroid based and learned prototype based performance on GZSL. VCB refers the visual centroid based method, which shares the same framework of VPB but with visual centroids instead of learned visual prototypes.}
 \label{fig:abl_proto_GZSL}
 \end{figure}

The comparison of VCB and VPB on GZSL is given in Fig. \ref{fig:abl_proto_GZSL}. In terms of harmonic accuracy, the visual prototype based method gets superior performance over the visual centroid based methods on all four datasets. Concretely, the VPB is better than the VCB by 14.5\%, 15.3\%, 8.9\%, and 10.4\% on AwA1, AwA2, CUB, and SUN respectively. It is worth noting that the VCB obtains better tr accuracy but obviously lower harmonic accuracy on the datasets AwA1, AwA2, and CUB than VPB, which indicates that the VCB tends to get overfitting results on the seen classes, whereas the proposed learned visual prototype based method shows outstanding performance on generalization.

\section{Discussion}
In terms of the overall performance on the ZSL and GZSL, the prototype based method achieves the most outstanding performance, especially on the task of GZSL. Compared with the direct mapping of the semantic vectors to the visual space, wherein the mapped semantic features need to be optimized according to numerous instance features of the same classes, the prototype based method makes each semantic vector have a clear mapping target, which is the visual prototype feature of the same class instead of a massive instance features. So that even the simple visual prototypes served by the visual centroids can achieve noticeable performance. As shown in Fig. \ref{fig:abl_proto_GZSL} and Table \ref{tab:GZSL}, the VCB outperforms all previous methods on the datasets of AwA1 and AwA2. However, taking the visual centroids as the visual prototypes tends to overfit on the seen classes, since the interacting distribution of instances features in the visual space brings the fact that the visual centroids may very close to some instance features from other classes. The proposed learnable visual prototypes are more distinctive, which can effectively alleviate the overfitting problem and show excellent performance on the task of GZSL.

The visual structure optimizing methods are not good as the visual prototype based method, but they still get conspicuous performance, especially on the ZSL, compared with other existing methods. In the sight of embedding loss, the bi-directional ranking loss pays more attention on making the matched semantic and visual feature pairs closer and non-matched pairs farther. Intuitively, the bi-directional ranking would obtain a better performance than that of the simple ranking loss. However, as listed in Table \ref{tab:comparsion}, there is only a slight improvement using the bi-directional ranking loss compared with the simple one. There may be two reasons for the non-significant advantage of the bi-directional ranking loss. One is that the visual structure optimizing loss constrains the embedding visual features to have a good data structure. Under this condition, extra constraints on the semantic features or visual features are not necessary. The other reason may be that on one hand the bi-directional ranking loss makes the embedding semantic features more discriminative, while on the other hand this discrimination may disturb the relation among different categories. This may be the reason that BR is slightly weaker than SR on the AwA2 dataset. Nevertheless, the bi-directional ranking loss shows obviously better performance on the more realistic task GZSL, as shown in Table \ref{tab:GZSL}.

\section{Conclusion}
In this paper, we explore the idea of optimizing visual space for ZSL recognition. To this end, we introduce two methods, one of which is called prototype based method and the other is the visual data structure optimization based method. In the former method, we learns a visual prototype for each visual class, so that the semantic vector can be mapped with the certain visual prototype rather than numerous visual features that discretely distribute in the visual space. In the latter method, accompanied with a embedding loss, the proposed visual structure optimizing loss can effectively improve the performance on the ZSL and GZSL. For the embedding loss, we consider two forms, including a simple ranking loss and a bi-directional ranking loss. When the proposed optimizing loss is added to the framework, both of the ranking loss functions show outstanding performance on the ZSL task. Extensive experiments on four zero-shot benchmarks demonstrate the superiority of our proposed models, and the proposed visual prototype based method outperforms all the previous methods, achieving the new state-of-the-art performance.

Considering the generality, in the current proposed visual prototype based method, we only use the visual features to learn the prototypes but ignore the information presented by the corresponding training semantic vectors, with which, in fact, we can further optimize the visual prototypes to make the visual space formed by visual prototypes have closer manifold with the semantic space, and that will be conducted in our future research.
\section*{Acknowledgment}

This work was supported in part by the National Natural Science Foundation of China under Grants 61573273 and 61603289, and in part by the  Fundamental Research
Funds for the Central Universities under Grant xzy022019052.

\ifCLASSOPTIONcaptionsoff
  \newpage
\fi

\bibliographystyle{IEEEtran}
\bibliography{Zero_shot_learning_TMM.bib}

\begin{thebibliography}{10}
\providecommand{\url}[1]{#1}
\csname url@samestyle\endcsname
\providecommand{\newblock}{\relax}
\providecommand{\bibinfo}[2]{#2}
\providecommand{\BIBentrySTDinterwordspacing}{\spaceskip=0pt\relax}
\providecommand{\BIBentryALTinterwordstretchfactor}{4}
\providecommand{\BIBentryALTinterwordspacing}{\spaceskip=\fontdimen2\font plus
\BIBentryALTinterwordstretchfactor\fontdimen3\font minus
  \fontdimen4\font\relax}
\providecommand{\BIBforeignlanguage}[2]{{%
\expandafter\ifx\csname l@#1\endcsname\relax
\typeout{** WARNING: IEEEtran.bst: No hyphenation pattern has been}%
\typeout{** loaded for the language `#1'. Using the pattern for}%
\typeout{** the default language instead.}%
\else
\language=\csname l@#1\endcsname
\fi
#2}}
\providecommand{\BIBdecl}{\relax}
\BIBdecl

\bibitem{Krizhevsky2012ImageNet}
A.~Krizhevsky, I.~Sutskever, and G.~E. Hinton, ``Imagenet classification with
  deep convolutional neural networks,'' in \emph{International Conference on
  Neural Information Processing Systems}, 2012, Conference Proceedings, pp.
  1097--1105.

\bibitem{Simonyan2014Very}
K.~Simonyan and A.~Zisserman, ``Very deep convolutional networks for
  large-scale image recognition,'' \emph{Computer Science}, 2014.

\bibitem{He2016Deep}
K.~He, X.~Zhang, S.~Ren, and J.~Sun, ``Deep residual learning for image
  recognition,'' in \emph{IEEE Conference on Computer Vision and Pattern
  Recognition}, 2016, Conference Proceedings, pp. 770--778.

\bibitem{Salakhutdinov2011Learning}
R.~Salakhutdinov, A.~Torralba, and J.~Tenenbaum, ``Learning to share visual
  appearance for multiclass object detection,'' in \emph{IEEE Conference on
  Computer Vision and Pattern Recognition}, 2011, Conference Proceedings.

\bibitem{Zhu2014Capturing}
X.~Zhu, D.~Anguelov, and D.~Ramanan, ``Capturing long-tail distributions of
  object subcategories,'' in \emph{IEEE Conference on Computer Vision and
  Pattern Recognition}, 2014, Conference Proceedings.

\bibitem{Lampert2009Learning}
C.~H. Lampert, H.~Nickisch, and S.~Harmeling, \emph{Learning to detect unseen
  object classes by between-class attribute transfer}, 2009.

\bibitem{Larochelle2008Zero}
H.~Larochelle, D.~Erhan, and Y.~Bengio, ``Zero-data learning of new tasks,'' in
  \emph{AAAI Conference on Artificial Intelligence}, 2008, Conference
  Proceedings.

\bibitem{Palatucci2009Zero}
M.~Palatucci, D.~Pomerleau, G.~E. Hinton, and T.~M. Mitchell, ``Zero-shot
  learning with semantic output codes,'' in \emph{International Conference on
  Neural Information Processing Systems}, 2009, Conference Proceedings, pp.
  1410--1418.

\bibitem{Farhadi2009Describing}
A.~Farhadi, I.~Endres, D.~Hoiem, and D.~Forsyth, ``Describing objects by their
  attributes,'' in \emph{IEEE Conference on Computer Vision and Pattern
  Recognition}, 2009, Conference Proceedings.

\bibitem{Farhadi2010Attribute}
A.~Farhadi, I.~Endres, and D.~Hoiem, ``Attribute-centric recognition for
  cross-category generalization,'' in \emph{IEEE Conference on Computer Vision
  and Pattern Recognition}, 2010, Conference Proceedings.

\bibitem{Vittorio2008Learning}
A.~Z. Vittorio~Ferrari, ``Learning visual attributes,'' \emph{Advances in
  Neural Information Processing Systems}, pp. 433--440, 2008.

\bibitem{Mikolov2013Distributed}
T.~Mikolov, I.~Sutskever, K.~Chen, G.~S. Corrado, and J.~Dean, ``Distributed
  representations of words and phrases and their compositionality,'' in
  \emph{Advances in neural information processing systems}, 2013, Conference
  Proceedings, pp. 3111--3119.

\bibitem{Lei2015Predicting}
J.~Lei~Ba, K.~Swersky, and S.~Fidler, ``Predicting deep zero-shot convolutional
  neural networks using textual descriptions,'' in \emph{IEEE International
  Conference on Computer Vision}, 2015, Conference Proceedings, pp. 4247--4255.

\bibitem{Elhoseiny2014Write}
M.~Elhoseiny, B.~Saleh, and A.~Elgammal, ``Write a classifier: Zero-shot
  learning using purely textual descriptions,'' in \emph{IEEE International
  Conference on Computer Vision}, 2014, Conference Proceedings.

\bibitem{Radovanovic2010Hubs}
N.~Radovanovic.M, \emph{Hubs in Space: Popular Nearest Neighbors in
  High-Dimensional Data.}, 2010.

\bibitem{Shigeto2015Joint}
Y.~Shigeto, I.~Suzuki, K.~Hara, M.~Shimbo, and Y.~Matsumoto, ``Ridge
  regression, hubness, and zero-shot learning,'' in \emph{Joint European
  Conference on Machine Learning and Knowledge Discovery in Databases}.\hskip
  1em plus 0.5em minus 0.4em\relax Springer, 2015, Conference Proceedings, pp.
  135--151.

\bibitem{Xian2017Zero}
Y.~Xian, C.~H. Lampert, B.~Schiele, and Z.~Akata, ``Zero-shot learning - a
  comprehensive evaluation of the good, the bad and the ugly,'' \emph{IEEE
  Transactions on Pattern Analysis \& Machine Intelligence}, vol.~PP, no.~99,
  pp. 1--1, 2017.

\bibitem{WelinderEtal2010}
P.~Welinder, S.~Branson, T.~Mita, C.~Wah, F.~Schroff, S.~Belongie, and
  P.~Perona, ``{Caltech-UCSD Birds 200},'' California Institute of Technology,
  Tech. Rep. CNS-TR-2010-001, 2010.

\bibitem{Xiao2010SUN}
J.~Xiao, J.~Hays, K.~A. Ehinger, A.~Oliva, and A.~Torralba, ``Sun database:
  Large-scale scene recognition from abbey to zoo,'' in \emph{IEEE Conference
  on Computer Vision and Pattern Recognition}, 2010, Conference Proceedings.

\bibitem{Pan2010A}
S.~J. Pan and Q.~Yang, ``A survey on transfer learning,'' \emph{IEEE
  Transactions on knowledge and data engineering}, vol.~22, no.~10, pp.
  1345--1359, 2010.

\bibitem{Weiss2016A}
K.~Weiss, T.~M. Khoshgoftaar, and D.~Wang, ``A survey of transfer learning,''
  \emph{Journal of Big Data}, vol.~3, no.~1, p.~9, 2016.

\bibitem{Lampert2014Attribute}
C.~H. Lampert, N.~Hannes, and H.~Stefan, ``Attribute-based classification for
  zero-shot visual object categorization,'' \emph{IEEE Transactions on Pattern
  Analysis \& Machine Intelligence}, vol.~36, no.~3, pp. 453--465, 2014.

\bibitem{Zhao2018Domain}
A.~Zhao, M.~Ding, J.~Guan, Z.~Lu, and J.~R. Wen, ``Domain-invariant projection
  learning for zero-shot recognition,'' 2018.

\bibitem{Liu2018Zero}
Y.~Liu, Q.~Gao, J.~Li, J.~Han, and L.~Shao, ``Zero shot learning via low-rank
  embedded semantic autoencoder,'' in \emph{International Joint Conference on
  Artificial Intelligence}, 2018, Conference Proceedings, pp. 2490--2496.

\bibitem{Annadani2018Preserving}
Y.~Annadani and S.~Biswas, ``Preserving semantic relations for zero-shot
  learning,'' in \emph{IEEE Conference on Computer Vision and Pattern
  Recognition}, 2018, Conference Proceedings, pp. 7603--7612.

\bibitem{Long2017Zero}
Y.~Long, L.~Liu, F.~Shen, L.~Shao, and X.~Li, ``Zero-shot learning using
  synthesised unseen visual data with diffusion regularisation,'' \emph{IEEE
  Transactions on Pattern Analysis \& Machine Intelligence}, vol.~PP, no.~99,
  pp. 1--1, 2017.

\bibitem{Jiang2018Learning}
H.~Jiang, R.~Wang, S.~Shan, and X.~Chen, ``Learning class prototypes via
  structure alignment for zero-shot recognition,'' 2018.

\bibitem{Liu2018Generalized}
S.~Liu, M.~Long, J.~Wang, and M.~I. Jordan, ``Generalized zero-shot learning
  with deep calibration network,'' in \emph{Advances in Neural Information
  Processing Systems}, 2018, Conference Proceedings, pp. 2009--2019.

\bibitem{Verma2018Generalized}
A.~M. V.~Kumar~Verma, G.~Arora and P.~Rai, ``Generalized zero-shot learning via
  synthesized examples,'' in \emph{IEEE Conference on Computer Vision and
  Pattern Recognition}, 2018, Conference Proceedings.

\bibitem{Mishra2018A}
A.~Mishra, S.~Krishna~Reddy, A.~Mittal, and H.~A. Murthy, ``A generative model
  for zero shot learning using conditional variational autoencoders,'' in
  \emph{IEEE Conference on Computer Vision and Pattern Recognition Workshops},
  2018, Conference Proceedings, pp. 2188--2196.

\bibitem{Xian2018Feature}
Y.~Xian, T.~Lorenz, B.~Schiele, and Z.~Akata, ``Feature generating networks for
  zero-shot learning,'' in \emph{IEEE conference on computer vision and pattern
  recognition}, 2018, Conference Proceedings, pp. 5542--5551.

\bibitem{Felix2018Multi}
R.~Felix, B.~G.~V. Kumar, I.~Reid, and G.~Carneiro, ``Multi-modal
  cycle-consistent generalized zero-shot learning,'' 2018.

\bibitem{xian2019f}
Y.~Xian, S.~Sharma, B.~Schiele, and Z.~Akata, ``f-vaegan-d2: A feature
  generating framework for any-shot learning,'' 2019.

\bibitem{Jie2018Transductive}
S.~Jie, C.~Shen, Y.~Yang, L.~Yang, and M.~Song, ``Transductive unbiased
  embedding for zero-shot learning,'' 2018.

\bibitem{Fu2015Transductive}
Y.~Fu, T.~M. Hospedales, T.~Xiang, and S.~Gong, ``Transductive multi-view
  zero-shot learning,'' \emph{IEEE Transactions on Pattern Analysis \& Machine
  Intelligence}, vol.~37, no.~11, pp. 2332--2345, 2015.

\bibitem{Guo2016Transductive}
X.~J. Y.~Guo, G.~Ding and J.~Wang, ``Transductive zero-shot recognition via
  shared model space learning,'' in \emph{Thirtieth AAAI Conference on
  Artificial Intelligence}, 2016, Conference Proceedings.

\bibitem{Xian2016Latent}
Y.~Xian, Z.~Akata, G.~Sharma, Q.~Nguyen, M.~Hein, and B.~Schiele, ``Latent
  embeddings for zero-shot classification,'' in \emph{IEEE Conference on
  Computer Vision and Pattern Recognition}, 2016, Conference Proceedings, pp.
  69--77.

\bibitem{Romera-Paredes2015An}
B.~Romera-Paredes and P.~Torr, ``An embarrassingly simple approach to zero-shot
  learning,'' in \emph{International Conference on Machine Learning}, 2015,
  Conference Proceedings, pp. 2152--2161.

\bibitem{Akata2015Evaluation}
Z.~Akata, S.~Reed, D.~Walter, H.~Lee, and B.~Schiele, ``Evaluation of output
  embeddings for fine-grained image classification,'' in \emph{IEEE Conference
  on Computer Vision and Pattern Recognition}, 2015, pp. 2927--2936.

\bibitem{Frome2013Devise}
A.~Frome, G.~S. Corrado, J.~Shlens, S.~Bengio, J.~Dean, and T.~Mikolov,
  ``Devise: A deep visual-semantic embedding model,'' in \emph{Advances in
  neural information processing systems}, 2013, Conference Proceedings, pp.
  2121--2129.

\bibitem{Zhang2017Learning}
L.~Zhang, T.~Xiang, and S.~Gong, ``Learning a deep embedding model for
  zero-shot learning,'' 2017.

\bibitem{Changpinyo2016Synthesized}
S.~Changpinyo, W.~L. Chao, B.~Gong, and F.~Sha, ``Synthesized classifiers for
  zero-shot learning,'' in \emph{IEEE Conference on Computer Vision and Pattern
  Recognition}, 2016, Conference Proceedings.

\bibitem{Zhang2015Zero}
Z.~Zhang and V.~Saligrama, ``Zero-shot learning via joint latent similarity
  embedding,'' \emph{Computer Science}, 2015.

\bibitem{Kodirov2017Semantic}
E.~Kodirov, T.~Xiang, and S.~Gong, ``Semantic autoencoder for zero-shot
  learning,'' 2017.

\bibitem{Morgado2017Semantically}
P.~Morgado and N.~Vasconcelos, ``Semantically consistent regularization for
  zero-shot recognition,'' in \emph{IEEE Conference on Computer Vision and
  Pattern Recognition}, 2017, Conference Proceedings, pp. 6060--6069.

\bibitem{Meng2018Meng}
M.~Meng and X.~Zhan, ``Zero-shot learning via low-rank-representation based
  manifold regularization,'' \emph{IEEE Signal Processing Letters}, vol.~25,
  no.~9, pp. 1379--1383, 2018.

\bibitem{Zhang2016Zero-r}
Z.~Zhang and V.~Saligrama, ``Zero-shot recognition via structured prediction,''
  in \emph{European conference on computer vision}.\hskip 1em plus 0.5em minus
  0.4em\relax Springer, 2016, Conference Proceedings, pp. 533--548.

\bibitem{Li2017Zero}
Y.~Li, D.~Wang, H.~Hu, Y.~Lin, and Y.~Zhuang, ``Zero-shot recognition using
  dual visual-semantic mapping paths,'' in \emph{IEEE Conference on Computer
  Vision and Pattern Recognition}, 2017, Conference Proceedings, pp.
  3279--3287.

\bibitem{Deutsch2017Zero}
S.~Deutsch, S.~Kolouri, K.~Kim, Y.~Owechko, and S.~Soatto, ``Zero shot learning
  via multi-scale manifold regularization,'' in \emph{IEEE Conference on
  Computer Vision and Pattern Recognition}, 2017, Conference Proceedings, pp.
  7112--7119.

\bibitem{Zhang2018Triple}
H.~Zhang, Y.~Long, Y.~Guan, and L.~Shao, ``Triple verification network for
  generalized zero-shot learning,'' \emph{IEEE Transactions on Image
  Processing}, vol.~28, no.~1, pp. 506--517, 2019.

\bibitem{Akata2013Label}
Z.~Akata, F.~Perronnin, Z.~Harchaoui, and C.~Schmid, ``Label-embedding for
  attribute-based classification,'' in \emph{IEEE Conference on Computer Vision
  and Pattern Recognition}, 2013, Conference Proceedings, pp. 819--826.

\bibitem{Russakovsky2015ImageNet}
O.~Russakovsky, J.~Deng, H.~Su, J.~Krause, S.~Satheesh, S.~Ma, Z.~Huang,
  A.~Karpathy, A.~Khosla, and M.~Bernstein, ``Imagenet large scale visual
  recognition challenge,'' \emph{International Journal of Computer Vision},
  vol. 115, no.~3, pp. 211--252, 2015.

\bibitem{Angeliki2014Hubness}
D.~G. L.~Angeliki and B.~Marco, ``Hubness and pollution: Delving into
  cross-space mapping for zero-shot learning,'' in \emph{the 7th International
  Joint Conference on Natural Language Processing}, 2014, Conference
  Proceedings.

\bibitem{Norouzi2013Zero}
M.~Norouzi, T.~Mikolov, S.~Bengio, Y.~Singer, J.~Shlens, A.~Frome, G.~S.
  Corrado, and J.~Dean, ``Zero-shot learning by convex combination of semantic
  embeddings,'' \emph{arXiv preprint arXiv:1312.5650}, 2013.

\bibitem{Socher2013Zero}
R.~Socher, M.~Ganjoo, H.~Sridhar, O.~Bastani, C.~D. Manning, and A.~Y. Ng,
  ``Zero-shot learning through cross-modal transfer,'' in \emph{International
  Conference on Neural Information Processing Systems}, 2013, Conference
  Proceedings.

\bibitem{Zhang2016Zero}
Z.~Zhang and V.~Saligrama, ``Zero-shot learning via joint semantic similarity
  embedding.''\hskip 1em plus 0.5em minus 0.4em\relax IEEE Conference on
  Computer Vision and Pattern Recognition, 2016, Conference Proceedings.

\bibitem{Akata2016Label}
Z.~Akata, F.~Perronnin, Z.~Harchaoui, and C.~Schmid, ``Label-embedding for
  image classification,'' \emph{IEEE Transactions on Pattern Analysis \&
  Machine Intelligence}, vol.~38, no.~7, pp. 1425--1438, 2016.

\bibitem{Verma2018Verma}
V.~K. Verma and P.~Rai, ``A simple exponential family framework for zero-shot
  learning,'' 2017.

\bibitem{ding2019marginalized}
Z.~Ding and H.~Liu, ``Marginalized latent semantic encoder for zero-shot
  learning,'' in \emph{Proceedings of the IEEE Conference on Computer Vision
  and Pattern Recognition}, 2019, pp. 6191--6199.

\bibitem{schonfeld2019generalized}
E.~Schonfeld, S.~Ebrahimi, S.~Sinha, T.~Darrell, and Z.~Akata, ``Generalized
  zero-and few-shot learning via aligned variational autoencoders,'' in
  \emph{Proceedings of the IEEE Conference on Computer Vision and Pattern
  Recognition}, 2019, pp. 8247--8255.

\bibitem{huang2019generative}
H.~Huang, C.~Wang, P.~S. Yu, and C.-D. Wang, ``Generative dual adversarial
  network for generalized zero-shot learning,'' in \emph{Proceedings of the
  IEEE Conference on Computer Vision and Pattern Recognition}, 2019, pp.
  801--810.

\bibitem{bulent2019gradient}
M.~Bulent~Sariyildiz and R.~Gokberk~Cinbis, ``Gradient matching generative
  networks for zero-shot learning,'' in \emph{Proceedings of the IEEE
  Conference on Computer Vision and Pattern Recognition}, 2019, pp. 2168--2178.

\bibitem{li2019leveraging}
J.~Li, M.~Jin, K.~Lu, Z.~Ding, L.~Zhu, and Z.~Huang, ``Leveraging the invariant
  side of generative zero-shot learning,'' 2019.

\bibitem{xie2019attentive}
G.-S. Xie, L.~Liu, X.~Jin, F.~Zhu, Z.~Zhang, J.~Qin, Y.~Yao, and L.~Shao,
  ``Attentive region embedding network for zero-shot learning,'' in
  \emph{Proceedings of the IEEE Conference on Computer Vision and Pattern
  Recognition}, 2019, pp. 9384--9393.

\end{thebibliography}
\end{document}